%% file: neurips_2025.tex
\documentclass{article}

\usepackage[final]{neurips_2025}

\usepackage{amsfonts}       
\usepackage{amsmath}
\usepackage{amssymb}
\usepackage{amsthm}
\usepackage{algorithm}
\usepackage{algorithmic}
\usepackage{booktabs}       
\usepackage{colortbl}
\usepackage{enumitem}
\usepackage[T1]{fontenc}    
\usepackage{hyperref}       
\usepackage[utf8]{inputenc} 
\usepackage{graphicx}
\usepackage{makecell}
\usepackage{mathtools}
\usepackage{microtype}      
\usepackage{multirow}
\usepackage{nicefrac}       
\usepackage{rotating}
\usepackage{subfigure}
\usepackage{url}            
\usepackage{wrapfig}
\usepackage{xcolor}         

\newcommand\mathrb[1]{\mathrm{\mathbf{#1}}}
\makeatletter
\newcommand\figcaption{\def\@captype{figure}\caption}
\newcommand\tabcaption{\def\@captype{table}\caption}
\makeatother

\usepackage{hyperref}
\definecolor{color1}{HTML}{006EB8}
\hypersetup{
  colorlinks   = true,
  urlcolor     = color1,
  linkcolor    = color1,
  citecolor   = color1
}

\newtheorem{defi}{Definition}[section]
\newtheorem{lem}[defi]{Lemma}%
\newtheorem{thrm}[defi]{Theorem}%

\title{Relieving the Over-Aggregating Effect in \\Graph Transformers}

\author{%
  Junshu Sun$^{1,2}$\hspace{1em}
  Wanxing Chang$^3$\hspace{1em}\and
  \textbf{Chenxue Yang}$^{4*}$\hspace{1em}
  \textbf{Qingming Huang}$^{1,2}$\hspace{1em}
  \textbf{Shuhui Wang}$^{1}$\thanks{Corresponding author.}\\
  $^1$Institute of Computing Technology, CAS\hspace{1em}
  $^2$University of Chinese Academy of Sciences\\
  $^3$DAMO Academy, Alibaba Group\hspace{1em}
  $^4$Agriculture Information Institute, CAAS\\
  \texttt{\{sunjunshu21s,wangshuhui\}@ict.ac.cn}\hspace{1em}
  \texttt{changwanxing.cwx@alibaba-inc.com} \\
  \texttt{yangchenxue@caas.cn
  qmhuang@ucas.ac.cn}\\ 
}

\begin{document}

\maketitle

\begin{abstract}
Graph attention has demonstrated superior performance in graph learning tasks. However, learning from global interactions can be challenging due to the large number of nodes. In this paper, we discover a new phenomenon termed over-aggregating. Over-aggregating arises when a large volume of messages is aggregated into a single node with less discrimination, leading to the dilution of the key messages and potential information loss. To address this, we propose Wideformer, a plug-and-play method for graph attention. Wideformer divides the aggregation of all nodes into parallel processes and guides the model to focus on specific subsets of these processes. The division can limit the input volume per aggregation, avoiding message dilution and reducing information loss. The guiding step sorts and weights the aggregation outputs, prioritizing the informative messages. Evaluations show that Wideformer can effectively mitigate over-aggregating. As a result, the backbone methods can focus on the informative messages, achieving superior performance compared to baseline methods. 
\end{abstract}

\section{Introduction}
Transformers have achieved remarkable success in modeling Euclidean-structured data, including natural language processing~\cite{vaswani_AttentionAllYou_2017,liu_DoRAWeightDecomposedLowRank_2024}, image understanding~\cite{liu_SwinTransformerHierarchical_2021a,shen_ExpandingSparseTuning_2024,shen_EnhancingPretrainedRepresentation_2024}, and video processing~\cite{arnab_ViViTVideoVision_2021,wu_VideoLanguageModel_2025}. Built on this success, significant efforts have been devoted to adapting transformers to graph-structured data, giving rise to various graph transformer models~\cite{li_GraphTransformer_2018,yun_GraphTransformerNetworks_2019,ying_TransformersReallyPerform_2021}. Leveraging the attention mechanism, graph transformers can learn from long-range dependencies and capture global features. These strengths provide effective solutions to address the challenges of over-smoothing~\cite{oono_GraphNeuralNetworks_2021} and over-squashing~\cite{alon_BottleneckGraphNeural_2021,topping_UnderstandingOversquashingBottlenecks_2021}, commonly encountered in traditional graph neural networks (GNNs)~\cite{kipf_SemiSupervisedClassificationGraph_2017,hamilton_InductiveRepresentationLearning_2017,velickovic_GraphAttentionNetworks_2018,sun_AllinARow_2023,gutteridge_DRewDynamicallyRewired_2023,sun_DynamicMessagePassing_2024}. Evaluations across diverse downstream tasks highlight the promising performance of graph transformers~\cite{ying_TransformersReallyPerform_2021,shirzad_ExphormerSparseTransformers_2023,deng_PolynormerPolynomialExpressiveGraph_2023}.

\input{fig-tex/teaser}

Nevertheless, the dense global attention in graph transformers incurs a space complexity of $O(n^2)$, where the memory consumption increases quadratically with the number of input nodes. This high complexity hinders graph transformers from scaling to large datasets. To enhance the scalability of graph transformers, two primary approaches have been proposed, including sparse attention~\cite{kreuzer_RethinkingGraphTransformers_2021,shirzad_ExphormerSparseTransformers_2023} and linear global attention~\cite{wu_NodeFormerScalableGraph_2022,wu_DIFFormerScalableGraph_2023,wu_SGFormerSimplifyingEmpowering_2023,deng_PolynormerPolynomialExpressiveGraph_2023}. Sparse methods reduce the density of dense global attention, which, in its original form, corresponds to a fully connected graph. The sparsified attention graph contains fewer edges, resulting in reduced memory overhead but also smaller receptive fields for each node. In contrast, linear methods preserve pairwise interactions between nodes and reduce complexity by eliminating the need for explicitly computing the dense attention.

However, learning from pairwise interactions is challenging due to the large number of graph nodes. Fig.~\ref{fig:teaser} illustrates the average attention entropy on different models. Attention entropy denotes the confidence of the nodes in discriminating informative messages during aggregation. Higher entropy indicates a more uniform distribution of attention scores and less confidence. From Fig.~\ref{fig:teaser-all}, we can see that linear methods tend to aggregate messages from nodes with higher attention entropy and similar attention scores. When the number of nodes increases in Fig.~\ref{fig:teaser-node}, the attention scores become more and more similar. As a result, a large volume of messages is aggregated into a single node with less discrimination, leading to potential information loss. We name this uniform aggregation phenomenon as over-aggregating. 

To tackle over-aggregating in graph transformers, we propose Wideformer, a plug-and-play method for graph attention. Wideformer divides the aggregation of all nodes into parallel processes and guides the model to focus on specific subsets of these processes. Specifically, the aggregation inputs are divided into distinct clusters, with each cluster aggregated separately. Instead of compressing messages from all nodes into a single target node, Wideformer limits the input volumes for each aggregation, thereby avoiding message dilution and reducing information loss. To ensure target nodes focus on the informative messages, the aggregation results are sorted and weighed based on the attention scores between the clusters and the target nodes. In contrast to the sparse attention methods that reduce the number of inputs for each target node, Wideformer maintains the global receptive field and distributes the inputs into multiple aggregation processes. By integrating with three popular linear methods GraphGPS~\cite{rampasek_RecipeGeneralPowerful_2022}, SGFormer~\cite{wu_SGFormerSimplifyingEmpowering_2023}, and Polynormer~\cite{deng_PolynormerPolynomialExpressiveGraph_2023}, Wideformer demonstrates effectiveness in tackling over-aggregating. It can consistently benefit the backbones to achieve superior performance over baselines. Codes are available at \url{https://github.com/sunjss/over-aggregating}.

The contribution of this paper can be summarized as follows:
\begin{itemize}[leftmargin=0.3in]
    \item We discover a new phenomenon in graph attention, termed over-aggregating, which aggregates a large volume of messages with less discrimination.
    \item We propose a novel method to relieve over-aggregating, named Wideformer, which divides the aggregation process and guides the model to focus on the informative messages.
    \item We demonstrate the effectiveness of Wideformer on thirteen real-world datasets, which consistently relieves over-aggregating and benefits backbones to achieve superior performance.
\end{itemize}

\section{Related Work}
GNNs perform message passing on input graphs, enabling nodes to learn from their connected neighbors and capture topological features~\cite{zhang_CompleteExpressivenessHierarchy_2023a}. Among the different message passing methods, GAT~\cite{velickovic_GraphAttentionNetworks_2018} first adopts the attention mechanism to aggregate messages. It utilizes attention scores between connected nodes to help central nodes filter out noisy neighbors and focus on the most informative ones. Following GAT, subsequent methods further adopt dense global attention to graphs~\cite{li_GraphTransformer_2018}. A primary challenge for these methods lies in how to effectively encode the graph structure, as global attention performs message passing between all pairs of nodes while neglecting the underlying graph structure. Several solutions have been proposed, including combining global attention with GNNs to simultaneously encode global features and graph structures~\cite{zhang_GraphBertOnlyAttention_2020,wu_RepresentingLongRangeContext_2021,mialon_GraphiTEncodingGraph_2021a}, learning positional or structural encodings as node features~\cite{zhang_GraphBertOnlyAttention_2020,dwivedi_GraphNeuralNetworks_2021,dwivedi_GeneralizationTransformerNetworks_2021,kreuzer_RethinkingGraphTransformers_2021}, and incorporating graph structures as an attention bias~\cite{ying_TransformersReallyPerform_2021}.

Another challenge arising from the dense global attention is a space complexity of $O(n^2)$. Efforts to address this challenge can be broadly categorized into two approaches: sparse attention and linear global attention. Sparse attention stems from the Big Bird model~\cite{zaheer_BigBirdTransformers_2020}, which utilizes random masks, sliding windows, and partial global masks to achieve sparsification in traditional transformer architectures. Built upon Big Bird, GraphGPS incorporates these sparse mechanisms into a comprehensive framework for constructing graph transformers~\cite{rampasek_RecipeGeneralPowerful_2022}. In addition to Big Bird, GAT, which employs input graphs as attention graphs, can also be considered a form of sparse attention. Exphormer~\cite{shirzad_ExphormerSparseTransformers_2023} and Spexphormer~\cite{shirzad_EvenSparserGraph_2024} further transform the input graphs into expander graphs, thus better approximating the fully connected graph. However, all of these sparsification methods shrink the receptive fields of each node, leading to a trade-off between model complexity and the ability to capture global information. Different from sparse attentions, Wideformer maintains the global receptive field while requiring linear computing complexity. 

In contrast to sparse attention, linear global attention focuses on directly simplifying the computation of dense attention. For example, GraphGPS~\cite{rampasek_RecipeGeneralPowerful_2022} and Nodeformer~\cite{wu_NodeFormerScalableGraph_2022} extend the linear attention method, Performer~\cite{choromanski_RethinkingAttentionPerformers_2020}, from traditional transformer architectures to graph transformers. Performer enables nearly unbiased estimations of dense attention without explicitly computing the attention matrix. In addition to approximation methods, several models have been proposed to model pairwise relations in linear time~\cite{wu_SGFormerSimplifyingEmpowering_2023,deng_PolynormerPolynomialExpressiveGraph_2023}. However, as shown in Fig.~\ref{fig:teaser-all}, the linear methods that maintain the global receptive field during aggregation encounter over-aggregating, leading to dilution of important messages and potential loss of information.

\section{Over-Aggregating in Graph Transformers}\label{sec:over-aggregating}
Graph transformers facilitate global or approximated global message passing between nodes. However, extracting informative features from these global interactions becomes challenging as the number of graph nodes increases. To investigate the behavior of global interactions in graph transformers, we utilize entropy as a measure of attention scores and examine how graph nodes distribute their attention in the receptive field.

\subsection{Attention Entropy}
In graph attention, input features are projected into query, key, and value features. The value features are aggregated based on the similarity between the query and key features. Let $\mathrb{X}\in\mathbb{R}^{n\times d}$ denote the input node features, where $n$ denotes the number of nodes and $d$ denotes the number of features. Node features are mapped to the query, key, and value features through $\mathrb{Q}=\mathrb{XW}_Q$, $\mathrb{K}=\mathrb{XW}_K$, and $\mathrb{V}=\mathrb{XW}_V$, respectively. Let $\mathrb{R}=\mathrb{QK}^\top$, we now have graph attention as
\begin{align}
    \mathrb{H}\label{eq:attention-ori}
    &=\boldsymbol{\alpha}\mathrb{V}, \quad
    \boldsymbol{\alpha}_{i,j}
    =\mathtt{softmax}(\mathrb{R})_{i,j}
    =\frac{\mathtt{exp}(\mathrb{R}_{i,j})}{\sum_k\mathtt{exp}(\mathrb{R}_{i,k})}.
\end{align}
Given the attention scores $\{\boldsymbol{\alpha}_{i,1}, \cdots, \boldsymbol{\alpha}_{i,n}\}$ between the target node $v_i$ and all graph nodes, the attention entropy can be formulated as
\begin{equation}
    \mathcal{H}(v_i)=-\sum_{j=1}^{n}
    \boldsymbol{\alpha}_{i,j}\log\boldsymbol{\alpha}_{i,j}.
\end{equation}
To ensure comparable results across different datasets, the entropy values are normalized by $\log~n$ to range $[0,1]$ for visualizations. Attention entropy measures the distribution of attention scores, indicating the confidence of the target nodes in discriminating informative messages. Lower entropy indicates that each target node focuses on a subset of graph nodes with higher attention scores (confidence), while higher entropy reflects more uniformly distributed attention and lower confidence. 

Fig.~\ref{fig:teaser-all} compares the attention entropy in sparse attention and linear global attention methods, including linear methods GraphGPS with Performer~\cite{rampasek_RecipeGeneralPowerful_2022}, Polynormer~\cite{deng_PolynormerPolynomialExpressiveGraph_2023}, and SGFormer~\cite{wu_SGFormerSimplifyingEmpowering_2023}, as well as sparse methods SAN~\cite{kreuzer_RethinkingGraphTransformers_2021} and Exphormer~\cite{shirzad_ExphormerSparseTransformers_2023}. The entropy values are averaged over all graph nodes throughout the training process. As shown in Fig.~\ref{fig:teaser-all}, linear and sparse methods exhibit distinct attention distribution patterns. Specifically, sparse methods demonstrate low attention entropy, which can be attributed to their sparsification strategies that enable nodes to filter out noisy messages and focus on the most informative ones. Conversely, linear methods involve all the graph nodes for aggregation and tend to assign similar attention scores in the receptive field, resulting in higher entropy. The distinct distribution patterns of the sparse and linear methods indicate the relation between the number of nodes in aggregation and the attention entropy.

\begin{thrm}[Monotonic Lower Bond of Attention Entropy]\label{thrm:mono-lower-bound}
Let $\boldsymbol{\alpha}_{i,\cdot}\in\mathbb{R}^{n}$ be the attention distribution of node $v_i$ over $n$ nodes. The attention entropy $\mathcal{H}(v_i)$ admits a lower bound that increases monotonically with $n$.
\end{thrm}

The proof is provided in Appendix~\ref{sec:app-proof}. Theorem~\ref{thrm:mono-lower-bound} indicates that graph attention will be more likely to have higher attention entropy when aggregating more graph nodes, which is further demonstrated in Fig.~\ref{fig:teaser-node}.

\subsection{Over-Aggregating}~\label{ssec:over-aggregating}
We refer to the aggregation process with high attention entropy as over-aggregating, typically demonstrated as uniform and small attention scores, offering limited discrimination among nodes. As a result, over-aggregating prevents graph nodes from prioritizing the informative messages, leading to message dilution and potential information loss.

\input{tab-tex/teaser}
To empirically evaluate the effect of over-aggregating with high attention entropy, we add an entropy regularization term to the optimization loss for linear methods GraphGPS with Performer, SGFormer, and Polynormer. The best results among the three backbones are reported. Tab.~\ref{tab:teaser} shows that directly reducing the attention entropy during training gives rise to better model performance. This indicates that over-aggregating with high attention entropy does harm the model performance. However, although directly regularizing attention entropy during training can benefit model performance, this process requires the explicit computation of the whole attention matrix, making it inapplicable to large graphs. As a result, it remains an open problem to alleviate over-aggregating for learning on graphs.

Over-aggregating stems from the optimization of attention scores. As shown in Fig.~\ref{fig:teaser-train}, models are typically initialized with high attention entropy, which gradually decreases during training. Throughout this process, models update the parameters $\mathrb{W}_Q$ and $\mathrb{W}_K$ with back-propagation, and thus optimize the attention scores. Take $\mathrb{W}_Q$ as an example, the gradient of $\boldsymbol{\alpha}_{i,j}$ with respect to $\mathrb{W}_Q$ can be derived as
\begin{equation}\label{eq:gradient}
    \begin{aligned}
        \frac{\partial\boldsymbol{\alpha}_{i,j}}{\partial\mathrb{W}_Q}
        \textcolor[RGB]{255,65,65}{\downarrow}
        &=\boldsymbol{\alpha}_{i,j}\textcolor[RGB]{255,65,65}{\downarrow}
        \!\left(
        \mathrb{X}_{i,\cdot}^\top\mathrb{X}_{j,\cdot}\mathrb{W}_K^\top-
        \!\sum_{\tiny k=1}^{n}\!\boldsymbol{\alpha}_{i,k}
        \mathrb{X}_{i,\cdot}^\top\mathrb{X}_{k,\cdot}\mathrb{W}_K^\top\right).
    \end{aligned}
\end{equation}
For more details, please refer to Appendix~\ref{sec:app-derivation}. A higher initial attention entropy implies smaller attention scores, which in turn attenuate the signals to update these scores. As a result, models converge with persistently high attention entropy, reflecting suboptimal attention distributions. Eq.~\ref{eq:gradient} can also provide insights for the aggravated over-aggregating problem on large-scale graphs (indicated by Theorem~\ref{thrm:mono-lower-bound} and Fig.~\ref{fig:teaser-node})
\begin{equation}
    \begin{aligned}
        \frac{\partial\boldsymbol{\alpha}_{i,j}}{\partial\mathrb{W}_Q}
        &=\boldsymbol{\alpha}_{i,j}(g_{v_j}-
        g_{\mathtt{others}}\textcolor[RGB]{255,65,65}{\uparrow})
        =\boldsymbol{\alpha}_{i,j}
        \!\left[
        (1-\boldsymbol{\alpha}_{i,j})\mathrb{X}_{i,\cdot}^\top\mathrb{X}_{j,\cdot}\mathrb{W}_K^\top-
        \!\sum_{\tiny \substack{k=1\\k\ne j}}^{n\textcolor[RGB]{255,65,65}{\uparrow}}
        \!\boldsymbol{\alpha}_{i,k}
        \mathrb{X}_{i,\cdot}^\top\mathrb{X}_{k,\cdot}\mathrb{W}_K^\top\right],
    \end{aligned}
\end{equation}
where $g_{v_j}$ corresponds to the update signals based on the interaction between node $v_i$ and $v_j$, and $g_{\mathtt{others}}$ corresponds to the cumulative impact of all other nodes. As the number of graph nodes grows, $g_{\mathtt{others}}$ increasingly incorporates more irrelevant signals, diluting the relative strength of the meaningful update signals in $g_{v_j}$. This dilution hinders the optimization of $\boldsymbol{\alpha}_{i,j}$, preventing the model from learning discriminative attention distributions and amplifying the over-aggregating issue on large-scale graphs.

\input{fig-tex/main}

\subsection{Discussion}
Similar to the occurrence of over-aggregating on the attention graphs, a recent study highlights a phenomenon known as over-globalizing in graph transformers, where models struggle to learn from neighboring nodes in homophilic datasets~\cite{xing_LessMoreGlobalizing_2024}. While over-globalizing specifically concerns nodes at varying distances, over-aggregating focuses on the discrimination of all nodes during aggregation. 

In the context of message passing on input graphs, over-dilution studies aggregation for both feature dimensions and neighboring nodes~\cite{lee_UnderstandingTacklingDilution_2025}. Specifically, intra-node dilution focuses on the volumes of the node features, while inter-node dilution focuses on whether a node's own information is preserved during aggregation. Different from over-dilution, over-aggregating focuses on the volumes of the input nodes and emphasizes the preservation of messages from the informative nodes, regardless of whether that includes the node itself.

Besides over-dilution, over-smoothing occurs when node features become increasingly similar as the number of layers grows~\cite{oono_GraphNeuralNetworks_2021}. This results from the low-pass aggregation among connected nodes in the input graph. In contrast, over-aggregating occurs in the attention graph due to the large number of nodes in the global receptive field. Designing high-pass aggregating functions can tackle over-smoothing. Conversely, for over-aggregating on graph transformers, even when a high-pass behavior is desired, the optimization of attention weights may still be hindered by the overwhelming number of nodes involved in aggregation.

Another notable issue on input graphs is over-squashing, which occurs when message pathways include edges with negative curvatures. It leads to an exponential expansion of the receptive field as the message-passing distance increases~\cite{alon_BottleneckGraphNeural_2021}. In consequence, a large volume of messages is compressed into fixed-dimensional representations, impeding the effective message exchange between long-range nodes. Unlike over-squashing that occurs during multi-step message passing, over-aggregating arises in one step. It aggregates a large volume of messages into a single node with limited discrimination, leading to the loss of informative messages from one-hop neighbors. 

\section{Wideformer: Relieving the Over-Aggregating}
Based on the analysis in Sec.~\ref{sec:over-aggregating}, global attention, including both dense and linear methods, faces a critical trade-off between mitigating over-aggregating and maintaining a global receptive field. To address this challenge, we propose Wideformer, a plug-and-play method designed to enhance graph transformer backbones. Unlike sparse attention methods that reduce the input volume, Wideformer maintains the total inputs and increases the output volume. Specifically, it partitions the inputs into distinct clusters, giving rise to multiple aggregation results for each node and guiding the nodes to focus on the informative clusters. The following subsections detail the two key components of Wideformer: dividing the aggregation and guiding the attention.

\subsection{Dividing the Aggregation}
Message passing involves aggregating messages from source nodes to target nodes. For each target node $v_i$ in global attention, it treats all the input nodes $\{v_j\},j\in\{1, \cdots,n\}$ as the source nodes, and aggregate messages $\mathrb{V}_{j,\cdot}$ based on the attention scores between its query features $\mathrb{Q}_{i,\cdot}$ and the source key features $\mathrb{K}_{j,\cdot}$. As a result, messages from all source nodes at each feature dimension are compressed into a single-dimensional representation, leading to the potential dilution of the key messages. 

\begin{wrapfigure}{r}{0.52\linewidth}
\vspace{-0.3in}
\resizebox{0.9\linewidth}{!}{
\begin{minipage}{\linewidth}
\begin{algorithm}[H]
\caption{Center-selection Function $\mathtt{Cluster}$}
\label{alg:center}
\begin{algorithmic}
\STATE {\bfseries Input:} query $\mathrb{Q}\in\mathbb{R}^{n\times d}$, 
number of clusters $m$
\STATE Initialize center $\mathrb{C}=\{0\}_{m\times d}$
\STATE $\mathrb{C}_{1,\cdot}=\mathrb{Q}_{t,\cdot},\ t=\arg\max_i\sum_j\mathrb{Q}_{i,j}$
\FOR{$t = 1$ {\bfseries to} $m$}
   \STATE Compute distances:
   $\mathrb{q} = \max_{j}\left[(\mathrb{Q}\mathrb{C}^\top)_{\cdot,j}\right]$
   \STATE Select next center indices: $q= \arg\min_i(\mathrb{q}_i)$
   \STATE Update centers: $\mathrb{C}_{t,\cdot}=\mathrb{Q}_{q,\cdot}$
\ENDFOR
\STATE {\bfseries Output:} centers $\mathrb{C}\in\mathbb{R}^{m\times d}$
\end{algorithmic}
\end{algorithm}
\end{minipage}}
\vspace{-0.1in}
\end{wrapfigure}
To relieve the dilution, we propose to divide the source nodes into different clusters based on their attention scores and aggregate each cluster separately. Therefore, source nodes with the highest level of attention can be aggregated together, excluding the other nodes and relieving the message dilution. However, explicitly computing the attention matrix requires a space complexity of $O(n^2)$. To address this issue, Wideformer approximates the division with attention scores by first selecting cluster centers among the query features, and then assigning the source nodes to the clusters based on the similarity between their key features and the centers. 

Let $m$ be the number of clusters, $\mathrb{C}=\mathtt{Cluster}(\mathrb{Q}, m)\in\mathbb{R}^{m\times d}$ be the cluster centers. The center-selection function $\mathtt{Cluter}$ in Algorithm~\ref{alg:center} is implemented based on K-Means++~\cite{arthur_KmeansAdvantagesCareful_2007}. The query of the source node with the largest summation across feature dimensions is selected as the initial cluster center. Notably, this initial center is solely used to initiate the selection process and does not participate in the aggregation. $\mathtt{Cluter}$ then select $m$ centers sequentially, where the node that minimizes the maximum similarity to all previously selected centers is designated as a new cluster center.

The source nodes are assigned to the identified query clusters based on the similarity between $\mathrb{K}$ and $\mathrb{C}$. The assignment result $\mathrb{k}\in\mathbb{R}^{n}$ can be formulated as
\begin{equation}
\mathrb{k}_i=
\arg\max_j\left[\left(\mathrb{KC}^\top\right)_{i,j}\right].
\end{equation}
By performing aggregation within distinct clusters, Wideformer produces multiple outputs for each target node. The aggregation result of the $t$-th cluster $\mathrb{H}^{(t)}\in\mathbb{R}^{n\times d}$ can be formulated as
\begin{equation}\label{eq:attention-wide}
\mathrb{H}^{(t)}
=\sum_{i\in\{j|\mathrb{k}_j=t\}}\mathtt{softmax}(\mathrb{Q}\mathrb{K}_{i,\cdot}^\top)\mathrb{V}_{i,\cdot}\ .
\end{equation}
To illustrate the effect of aggregation, for each target node $v_i$ and the $j$-th feature dimension of the source message $\mathrb{V}$, the original aggregation method in Eq.~\ref{eq:attention-ori} compresses messages from all nodes into a one-dimensional representation $\mathrb{H}_{i,j}$, which leads to the dilution of key messages and over-aggregating. In contrast, Wideformer with Eq.~\ref{eq:attention-wide} increases the output volume, yielding a $m$-dimensional representation $\{\mathrb{H}^{(1)}_{i,j},\cdots,\mathrb{H}^{(m)}_{i,j}\}$. Only a limited volume of messages is aggregated for each cluster, preventing the dilution of the key messages. Even when the target nodes apply uniform attention scores to the source nodes, the messages in different clusters remain separate from being over-aggregated.

\input{tab-tex/cmp-hom}
\subsection{Guiding the Attention}
The aggregation results $\{\mathrb{H}^{(1)},\cdots,\mathrb{H}^{(m)}\}$ are concatenated for the subsequent modules in backbones. During the concatenation, the current ordering of the aggregated results depends on the cluster-selection order. However, since each target node assigns different attention scores to the clusters, this order does not reflect the importance of each cluster for the target node. To ensure consistent ordering and enable the target nodes to focus on the informative clusters, Wideformer sorts and weights the clusters for each target node based on the cluster attention score $\bar{\boldsymbol{\alpha}}_{i,j}\in\mathbb{R}^{n\times m}$. These scores can be formulated as
\begin{equation}\label{eq:cluster-att}
\begin{aligned}
    &\bar{\boldsymbol{\alpha}}_{i,j}=\mathtt{softmax}(\mathrb{Q}\bar{\mathrb{K}}^\top)_{i,j},\quad
    &\bar{\mathrb{K}}_{t,\cdot}=
    \frac{1}{|\{v_i|\mathrb{k}_i=t\}|}
    \sum_{i\in\{j|\mathrb{k}_j=t\}}\mathrb{K}_{i,\cdot},
\end{aligned}
\end{equation}
where $\bar{\mathrb{K}}\in\mathbb{R}^{m\times d}$ denotes the average key features of the clusters. The sorted and weighted aggregation result $\hat{\mathrb{H}}^{(t)}\in\mathbb{R}^{n\times d}$ is then obtained as
\begin{equation}\label{eq:sort-weight}
\begin{aligned}
&\hat{\mathrb{H}}^{(t)}_{i,\cdot}
=\bar{\boldsymbol{\alpha}}_{i,\mathrb{S}_{i,t}}
\mathrb{H}^{(\mathrb{S}_{i,t})}_{i,\cdot},\quad
&\mathrb{S}_{i,\cdot}=
\arg\mathtt{sort}_j\left[\bar{\boldsymbol{\alpha}}_{i,j}\right],
\end{aligned}
\end{equation}
where $\arg\mathtt{sort}$ returns the indices of the sorted elements in ascending order. Eq.~\ref{eq:cluster-att} and \ref{eq:sort-weight} inject the importance of each cluster into the aggregation, guiding the focus of the target nodes. The weighted results are concatenated for the following computation.

\input{fig-tex/over-agg}

\section{Experiment}
To evaluate the effectiveness of Wideformer, we conduct empirical evaluations on real-world datasets. The detailed setups are presented in Appendix~\ref{sec:app-exp}.

\subsection{Effectiveness in tackling Over-Aggregating}
Wideformer is evaluated on three linear methods, including GraphGPS with Performer~\cite{rampasek_RecipeGeneralPowerful_2022}, SGFormer~\cite{wu_SGFormerSimplifyingEmpowering_2023}, and Polynormer~\cite{deng_PolynormerPolynomialExpressiveGraph_2023}. To ensure consistent comparison, these methods are implemented under the framework of GraphGPS. As presented in Fig.~\ref{fig:tackle-oag-all}, Wideformer can effectively reduce the attention entropy of all backbone methods. This reduction allows target nodes to focus on informative source nodes within the global receptive field, effectively preventing the dilution of key messages and mitigating over-aggregating.

\input{tab-tex/cmp-het}
However, Wideformer without the attention-guiding process still exhibits high attention entropy in Fig.~\ref{fig:tackle-oag-all}. This partial method only avoids message dilution but fails to effectively guide the attention of the target nodes. To further investigate the attention-guiding process in Wideformer, we conduct an empirical analysis of cluster attention $\bar{\boldsymbol{\alpha}}$. As shown in Fig.~\ref{fig:tackle-oag-n}, Wideformer provides low cluster attention entropy for all three backbones, indicating that only a subset of input messages is crucial for the target nodes. The attention-guiding process in Wideformer allows backbone methods to concentrate on these informative subsets, enhancing their ability to mitigate over-aggregating. As the number of nodes increases, GraphGPS and SGFormer exhibit increasingly divergent attention scores across clusters, while Polynormer maintains much lower cluster attention entropy, albeit with a slight upward trend. This indicates that Wideformer consistently supports different backbone methods in prioritizing informative messages, even when the graph size grows.

\subsection{Model Comparison}
\paragraph{Experimental Setup.}
We adopt thirteen real-world datasets, including both heterophilic and homophilic graphs. For baseline methods, both GNNs and graph transformers are adopted. Please refer to Appendix~\ref{ssec:app-setup} for detailed experimental setups. To address the out-of-memory issue associated with the entropy regularization baseline, we divide the target nodes into mini-batches. Each batch is sequentially processed to compute attention scores with all source nodes. The results of the backbone methods GraphGPS with Performer, SGFormer, and Polynormer are reproduced under the framework of GraphGPS.

\input{tab-tex/mix-large+ratio}
\paragraph{Performance.}
We present the comparison results on datasets where the number of nodes is around $10$k in Tab.~\ref{tab:cmp-hom} and Tab.~\ref{tab:cmp-het}. Wideformer consistently enhances the performance of all backbone models across various datasets. Specifically, backbones integrated with Wideformer achieve superior performance compared to baseline methods. For instance, in Tab.~\ref{tab:cmp-hom}, Wideformer enables backbone methods to outperform the previous best-performing baselines, including DIFFormer on AmazonComputers and NAGphormer on CoauthorsCS, CoauthorPhysics, and AmazonPhoto. Compared to the entropy regularization method, we can see that regularizing entropy results in inferior performance to Wideformer and even marginal gains on larger graphs. This is consistent with our analysis in Sec.~\ref{ssec:over-aggregating}. Entropy regularization does not divide the aggregation process, and thus suffers from optimization challenges when aggregating over a large number of source nodes. In contrast, Wideformer separates the aggregation into parallel processes. Each process only involves a small ratio of the source nodes to be aggregated, preventing the dilution of the key messages. We also compare the gains with zero to evaluate the statistical significance of the performance gains. Paired t-tests on all backbones consistently give rise to p-values smaller than 0.05, and the confidence interval results do not contain 0, validating the statistical significance of the performance gains with Wideformer. These results demonstrate that relieving over-aggregating improves the effectiveness of graph attention.

To further verify Wideformer's ability to mitigate over-aggregating, we extend our experiments to datasets with broader node scales, including $1$k and $0.1$M nodes. Results in Tab.~\ref{tab:cmp-mix} show that Wideformer can benefit backbones with different scales of nodes. Among the three backbones, SGFormer achieves better results on large-scale graphs ogb-arxiv and twitch-gamer, but inferior results on small-scale graphs CiteSeer and Cora. This is due to the trade-off between noise filtering and information loss in SGFormer, which is further investigated in Sec.~\ref{sssec:ana-source}.

\input{fig-tex/ana-node}

\subsection{Model Analysis}
\subsubsection{Informative Source Nodes}\label{sssec:ana-source}
Wideformer empowers backbone methods to focus on informative source nodes within the global receptive field. To investigate the attention mechanism in Wideformer, we conduct additional experiments on source nodes assigned to the cluster with the highest attention score, analyzing both their volumes and their relations with the target nodes.

\paragraph{Node Ratio.}
To analyze the node volume, we calculate the ratio of nodes assigned to the cluster with the highest attention score relative to the total number of graph nodes. As shown in Fig.~\ref{fig:att-node-cnt}, the node ratio decreases as the total number of nodes increases. This suggests that the proportion of informative nodes diminishes in larger graphs, leading to noisier source messages. Wideformer addresses this by assigning a smaller ratio of nodes to the cluster with the highest attention score, guiding the backbones to focus on the most informative nodes. 
Among the three backbones, SGFormer focuses on fewer source nodes than GraphGPS and Polynormer. This results in a trade-off between noise filtering and information loss, where SGFormer performs better on large-scale graphs (ogb-arxiv and twitch-gamer) but lags behind on smaller graphs (CiteSeer and Cora).

\paragraph{Node Relations.}
To analyze the relations between target and source nodes in the cluster with the highest attention score, we compute the ratio of the source nodes (1) that have the same label as the target nodes, and (2) that are connected with the target nodes, both relative to the total number of source nodes in the cluster. The results with GraphGPS are presented in Fig.~\ref{fig:rel-gps}. For the other two backbones, please refer to Appendix~\ref{sec:app-full-rel}. As shown in Fig.~\ref{fig:rel-gps}, we can see that approximately $10\%$ of source nodes are connected with their target nodes. The label distribution diverges between heterophilic graphs and homophilic graphs, where target nodes assign larger attention scores to the same-class source nodes on heterophilic graphs.

\subsubsection{Complexity Analysis}
The time and space complexity of Wideformer is $O(nm)$, where $n$ and $m$ denote the number of nodes and clusters, respectively. In practice, $m$ is chosen from $\{2, ..., 8\}$, which ensures the model's scalability to large graphs. We provide a time cost study regarding the number of nodes ($n$) and clusters ($m$), comparing the whole model with the part of Wideformer. The results on GraphGPS are presented in Fig.~\ref{fig:time-gps}, where the time cost of Wideformer scales linearly to $n$ and remains a small ratio of the total cost. For detailed experimental setup and full results, please refer to Appendix~\ref{sec:app-time}.

\subsubsection{Model Ablation}\label{sssec:ablation}
To evaluate the contribution of the aggregation-dividing and attention-guiding process in Wideformer, we conduct an ablation study on the three backbone methods.

\input{fig-tex/abl-cluster}

\paragraph{Cluster Ablation.}
Wideformer partitions the source nodes into distinct clusters. To examine the impact of this division, we perform an ablation study on the number of clusters $m$ regarding performance gains. The gain is defined as $\frac{p_m - p_1}{p_1}$, where $p_1$ and $p_m$ denote the performance with a single cluster and $m$ clusters, respectively. The average performance gains across various datasets for all three backbones are shown in Fig.~\ref{fig:abl-cluster}. We can see that increasing the number of clusters initially increases the performance gain. However, when $m$ exceeds $6$, the performance gain starts to decline. This is because a relatively larger number of clusters can effectively reduce the input message volume, mitigating the dilution of key information. Nevertheless, excessively partitioning the source nodes disperses critical information into multiple clusters. This leads to the distracted attention of the target nodes, reducing the values of attention scores and increasing attention entropy. In practice, the optimal number of clusters is around $3\sim5$.

\input{tab-tex/abl-center}
\paragraph{Center Selection.}
To ensure computing efficiency, Wideformer employs a non-iterative strategy for center selection. We now compare this simple choice with (1) multi-step iterative clustering and (2) parameterized centers, to shed light on the future directions. 
Tab.~\ref{tab:abl-center} shows that applying iterative clustering can achieve better results, indicating the efficiency-performance trade-offs. Further modeling centers as parameters can address this trade-off, which does not require iterations and achieves the best results.

\input{tab-tex/abl-module}
\paragraph{Module Ablation.}
Tab.~\ref{tab:abl-module} shows that combining divided aggregation and attention guidance generally achieves better performance. However, using only divided aggregation can also benefit the backbones, such as on CoauthorCS and AmazonPhoto with Polynormer. This is likely due to the simple attention-guiding strategy, which directly determines the importance of each cluster as a whole. Under circumstances where fewer source nodes contain informative messages, this simple strategy leads to limited benefit capability. Exploring alternative guiding strategies will be the focus of our future work.

\section{Conclusion}
In this paper, we discovered a phenomenon in graph attention, termed over-aggregating. Attention with over-aggregating assigns uniformly distributed attention scores to the messages. As a result, the key messages are diluted during aggregation, leading to potential information loss. To relieve this problem, we proposed a novel plug-and-play method, Wideformer. Wideformer divides the source nodes for aggregation into distinct clusters and aggregates each cluster separately. To ensure focus on the informative source nodes, Wideformer applies attention guidance to sort and weight the clusters. Experiments on thirteen datasets show that Wideformer can consistently enhance the performance of backbones and effectively relieve over-aggregating. Please refer to Appendix.~\ref{sec:app-limitation} for the limitation discussion.

\begin{ack}
This work was supported in part by the National Key R$\&$D Program of China under Grant 2023YFC2508704, in part by the National Natural Science Foundation of China 62236008, and in part by the Fundamental Research Funds for the Central Universities. The authors would like to thank the anonymous reviewers for their helpful comments and suggestions that improved this manuscript.
\end{ack}

\bibliography{bibliography}
\bibliographystyle{plain}

\newpage
\appendix
\renewcommand{\thefigure}{S\arabic{figure}}
\renewcommand{\thetable}{S\arabic{table}}
\renewcommand{\theequation}{S\arabic{equation}}
\section{Proof for Monotonic Lower Bound of Attention Entropy}\label{sec:app-proof}
\begin{lem}[Lower Bound of Attention Entropy]\label{thrm:app-lower-bound}
Let $\boldsymbol{\alpha}_{i,\cdot} \in \mathbb{R}^n$ be the attention distribution of node $v_i$ over $n$ nodes. Suppose each entry satisfies $\boldsymbol{\alpha}_{i,j} \ge \epsilon > 0$. Then the attention entropy $\mathcal{H}(v_i)$ admits the following strict lower bound:
\[
\mathcal{H}(v_i) \ge -\left[1-(n-1)\epsilon\right] \log\left[1-(n-1)\epsilon\right] - (n-1)\epsilon \log \epsilon.
\]
\end{lem}

\begin{proof}
Assume $\epsilon > 0$ is a lower bound for all elements of $\boldsymbol{\alpha}_{i,\cdot}$. The minimum entropy under this constraint is attained when one entry takes the maximum possible weight and the remaining $(n-1)$ entries equal $\epsilon$, \textit{i.e.},
\[
\hat{\boldsymbol{\alpha}}_{i,\cdot} = \left(1-(n-1)\epsilon, \epsilon, \dots, \epsilon\right).
\]
The attention entropy of this distribution is
\[
\mathcal{H}_{\min}(v_i) = -\left[1-(n-1)\epsilon\right] \log\left[1-(n-1)\epsilon\right] - (n-1)\epsilon \log \epsilon.
\]
Since the Shannon entropy is strictly concave, any deviation from this distribution increases the entropy. Therefore, for any valid distribution $\boldsymbol{\alpha}_{i,\cdot}$ with elements bounded below by $\epsilon$, it holds that
\[
\mathcal{H}(v_i) \ge \mathcal{H}_{\min}(v_i).
\]
\end{proof}

We now provide the proof for Theorem~\ref{thrm:mono-lower-bound} based on Lemma~\ref{thrm:app-lower-bound}.

\textbf{Theorem~\ref{thrm:mono-lower-bound}} (Monotonic Lower Bound of Attention Entropy). \textit{Let $\boldsymbol{\alpha}_{i,\cdot}\in\mathbb{R}^{n}$ be the attention distribution of node $v_i$ over $n$ nodes. The attention entropy $\mathcal{H}(v_i)$ admits a lower bound that increases monotonically with $n$.}

\begin{proof}
To show that $\mathcal{H}_{\min}(v_i)$ increases monotonically with $n$, we compute its derivative with respect to $n$
\begin{equation*}
\begin{aligned}
    \frac{d\mathcal{H}_{\min}(v_i)}{dn}
    &=\epsilon\log\left[1-(n-1)\epsilon\right]+\left[1-(n-1)\epsilon\right]\frac{\epsilon}{\left[1-(n-1)\epsilon\right]}-\epsilon\log\epsilon\\
    &=\epsilon\log\left[1-(n-1)\epsilon\right]+\epsilon-\epsilon\log\epsilon\\
    &=\epsilon\log\frac{1-(n-1)\epsilon}{\epsilon}+\epsilon.
\end{aligned}
\end{equation*}
Since $\hat{\boldsymbol{\alpha}}_{i,\cdot}=(1-(n-1)\epsilon, \epsilon, \cdots, \epsilon)$ denoting the most non-uniform distribution $\boldsymbol{\alpha}_{i,\cdot}$ requires $0 < \epsilon \le1/n$, it follows that $1 - (n-1)\epsilon \ge \epsilon$, and thus
\[
\frac{d\mathcal{H}_{\min}(v_i)}{dn} > 0.
\]
This confirms that $\mathcal{H}_{\min}(v_i)$ increases monotonically with $n$.
\end{proof}

\section{Detailed Derivation}\label{sec:app-derivation}
In this section, we provide a detailed derivation for Eq.~\ref{eq:gradient}. Let $\mathrb{R}=\mathrb{QK}^\top$. Given the attention scores formulated in Eq.~\ref{eq:attention-ori}, the gradient of $\boldsymbol{\alpha}_{i,j}$ with respect to $\mathrb{W}_Q$ can be derived as
\begin{equation*}
\begin{aligned}
    \frac{\partial\boldsymbol{\alpha}_{i,j}}{\partial\mathrb{W}_Q}
    &=\sum_{k=1}^n\frac{\partial\boldsymbol{\alpha}_{i,j}}{\partial\mathrb{R}_{i,k}}\frac{\partial\mathrb{R}_{i,k}}{\mathrb{W}_Q}\\
    &=\sum_{k=1}^n\boldsymbol{\alpha}_{i,j}\left(\delta_{j,k}-\boldsymbol{\alpha}_{i,k}\right)\frac{\partial}{\partial\mathrb{W}_Q}\mathrb{X}_{i,\cdot}\mathrb{W}_Q\mathrb{W}_K^\top\mathrb{X}_{k,\cdot}^\top\\
    &=\sum_{k=1}^n\boldsymbol{\alpha}_{i,j}\left(\delta_{j,k}-\boldsymbol{\alpha}_{i,k}\right)\mathrb{X}_{i,\cdot}^\top\mathrb{X}_{k,\cdot}\mathrb{W}_K^\top\\
    &=\boldsymbol{\alpha}_{i,j}
    \!\left(
    \mathrb{X}_{i,\cdot}^\top\mathrb{X}_{j,\cdot}\mathrb{W}_K^\top-
    \!\sum_{\tiny k=1}^{n}\!\boldsymbol{\alpha}_{i,k}
    \mathrb{X}_{i,\cdot}^\top\mathrb{X}_{k,\cdot}\mathrb{W}_K^\top\right),
\end{aligned}
\end{equation*}
where $\delta_{j,k}$ is a sign function. When $k=j$, $\delta_{j,k}=1$, otherwise $\delta_{j,k}=0$.

\input{tab-tex/app-reg}
\section{Over-Aggregating Measurement}
In this paper, we propose to employ entropy to quantify over-aggregating. This stems from the observation that the attention scores distribute uniformly among different source nodes. To quantify the uniformity of the distribution, we adopt entropy as the measure following the common practice in the community~\cite{ying_HierarchicalGraphRepresentation_2018, chiang_ClusterGCNEfficientAlgorithm_2019, platonov_CriticalLookEvaluation_2023}. Except for entropy, other potential metrics include standard deviation (STD), Kullback-Leibler divergence (KLD), and Jensen-Shannon divergence (JSD). STD focuses on the dispersion of values. When the total number of values is large, but the number of dispersed values is small, the STD may remain small despite the distribution being uneven. Therefore, using STD to assess uniformity is not an accurate approach. KLD and JSD are used to assess the similarity between two distributions. It can measure the uniformity of a distribution by setting the reference distribution to be uniform. However, when the reference distribution is uniform, KLD equals the difference between the maximum entropy and the entropy of the given distribution, while JSD can be viewed as a combination of entropy and KLD. Therefore, we opted for the simpler and more commonly used entropy to directly measure the uniformity. 

Entropy also demonstrates effectiveness in benefiting model performance. In addition to the entropy regularization method on the whole attention matrix (Tab.~\ref{tab:teaser}), we compute the entropy of the cluster attention scores in Wideformer and employ the results as part of the optimization target. Results on roman-empire and CoauthorCS in Tab.~\ref{tab:app-reg} show that optimizing the cluster attention entropy for backbones with Wideformer can also benefit the model performance. This also indicates that the value of the attention entropy properly quantifies over-aggregating.

In Fig.~\ref{fig:tackle-oag-all}, a more significant entropy reduction occurs on Polynormer, but its associated performance gains are not superior to those of the other two backbone architectures. This is due to the distinct architectural design of the backbones. Different architectures inherently prioritize different attention patterns for optimal representation learning. While high attention entropy consistently indicates over-aggregating, the optimal entropy level varies across models. Thus, although Polynormer exhibits the most significant entropy reduction, this may still fall short of what is needed for its optimal performance.

\section{Details on Experiments}\label{sec:app-exp}
\subsection{Datasets}\label{ssec:app-dataset}
We adopt thirteen real-world datasets, including heterophilic graphs (amazon-ratings, minesweeper, questions, roman-empire, tolokers~\cite{platonov_CriticalLookEvaluation_2023}, and twitch-gamer~\cite{lim_LargeScaleLearning_2021}) and homophilic graphs (Cora, CiteSeer~\cite{sen_CollectiveClassificationNetwork_2008}, CoauthorCS, CoauthorPhysics, AmazonComputers, AmazonPhoto~\cite{shchur_PitfallsGraphNeural_2019}, and ogb-arxiv~\cite{hu_OpenGraphBenchmark_2020}).

\textbf{Questions} is derived from Yandex Q, spanning user activity from September 2021 to August 2022. Nodes represent users interested in ``medicine" and edges indicate answers to others' questions. The binary classification task predicts whether users remained active without account deletion or blocking. Node features are averaged FastText embeddings of profile descriptions.

\textbf{Amazon-ratings} is based on the Amazon product co-purchase network from the SNAP Datasets~\cite{snapnets}. Nodes represent products, and edges capture frequent co-purchase relationships. The task is to predict the average reviewer rating for each product, grouped into five ordinal classes. Node features are the average FastText embeddings of product descriptions.

\textbf{Tolokers} represents crowdsourcing participation data from the Toloka platform. Nodes correspond to contributors (``tolokers") active in at least one of 13 projects, with edges connecting those who completed the same tasks. The binary classification task predicts whether tolokers were banned from projects. Node features include profile attributes and task performance statistics.

\textbf{Minesweeper} is a synthetic 100x100 grid graph where nodes represent grid cells, with 20\% randomly assigned as mines. The task is to classify nodes as mines or non-mines. Node features are one-hot vectors encoding the count of neighboring mines, but 50\% of nodes have their features reset to unknown values, marked by a binary indicator.

\textbf{Twitch-gamer} is collected from the Twitch website with nodes representing users. Only users with mutual relations are incorporated in the dataset. The task is to predict whether the user accounts contain explicit content.

\textbf{Cora and CiteSeer} are citation graphs. In the Cora dataset, machine learning papers are grouped into seven distinct classes, while the CiteSeer dataset categorizes papers into six classes. Node features are derived from high-frequency words appearing in the content of the papers.

\textbf{CoauthorCS and CoauthorPhysics} are derived from the Microsoft Academic Graph. These datasets represent co-authorship networks, where nodes correspond to researchers, and edges signify their collaborative relationships. The features of each node reflect the frequency of keywords extracted from the publications of the respective authors. Graph labels identify the primary research domain of each author, distinguishing between computer science and physics.

\textbf{AmazonComputers and AmazonPhoto} are co-purchase networks constructed from Amazon. In these graphs, nodes represent products, while edges indicate co-purchase relationships between item pairs. Node features are derived from encoded customer review texts associated with each product, and graph labels classify the products into specific categories.  

\textbf{OGB-arXiv} is a citation graph composed of Computer Science papers from arXiv. In this dataset, nodes correspond to individual articles, and directed edges represent citation relationships. Node features are computed as the average of 128-dimensional word embeddings extracted from the title and abstract of each paper. The task involves predicting the primary category of arXiv articles among 40 possible classes.

\subsection{Experimental Setup}\label{ssec:app-setup}
The attention entropy experiments in Fig.~\ref{fig:teaser}, Fig.~\ref{fig:tackle-oag}, and Fig.~\ref{fig:att-node-cnt} are conducted on Cora, AmazonPhoto, AmazonComputers, CoauthorCS, CoauthorPhysics, tolokers, amazon-ratings, minesweeper, and ogb-arxiv. The cluster ablation study in Fig.~\ref{fig:abl-cluster} is conducted and averaged on all the datasets included in this paper. The number of clusters is $4$ in Fig.~\ref{fig:tackle-oag}, Fig.~\ref{fig:att-node-cnt}-\ref{fig:rel-gps}, Fig.~\ref{fig:app-rel}, and Tab.~\ref{tab:abl-center}-\ref{tab:abl-module}. Other experiments choose $m$ from $\{2,\cdots,8\}$.
For baseline methods, both GNNs (GCN~\cite{kipf_SemiSupervisedClassificationGraph_2017}, GraphSAGE~\cite{hamilton_InductiveRepresentationLearning_2017}, GAT~\cite{velickovic_GraphAttentionNetworks_2018}, GPRGNN~\cite{chien_AdaptiveUniversalGeneralized_2022}) and graph transformers (NAGphormer~\cite{chen_NAGphormerTokenizedGraph_2023}, NodeFormer~\cite{wu_NodeFormerScalableGraph_2022}, DIFFormer~\cite{wu_DIFFormerScalableGraph_2023}, GOAT~\cite{kong_GOATGlobalTransformer_2023}) are included. The results of the backbone methods, including GraphGPS with Performer, SGFormer, and Polynormer, are reproduced under the framework of GraphGPS. The framework is implemented with PyTorch~\cite{paszke_PyTorchImperativeStyle_2019} and PyTorch Geometric~\cite{fey_FastGraphRepresentation_2019}, and trained on a single NVIDIA A100. We perform grid search based on the validation performance of the models as follows:

\paragraph{GraphGPS.} We search the number of graph transformer layers in $\{1, \cdots, 6\}$, the number of hidden dimensions in $\{64, 80, 128, 256\}$, the number of heads in $\{1, 2, 4\}$, dropout in $\{0.1, 0.2, 0.3, 0.5\}$, and learning rate in $\{5e-4, 1e-3, 1e-2\}$. The rest hyperparameters are fixed as in the original implementation.

\paragraph{SGFormer.} The GNN backbone is implemented as GCN~\cite{kipf_SemiSupervisedClassificationGraph_2017}. The number of GCN layers is searched in $\{1, \cdots, 10\}$, the number of hidden dimensions in $\{64, 80, 128, 256\}$, the number of heads in $\{1, 2, 4\}$, dropout in $\{0.1, 0.2, 0.3, 0.5\}$, and learning rate in $\{1e-3, 5e-3, 1e-2\}$. The rest hyperparameters are fixed as in the original implementation.

\paragraph{Polynormer.} The GNN backbone is implemented as GAT~\cite{velickovic_GraphAttentionNetworks_2018}. We search the number of graph transformer layers in $\{1, \cdots, 6\}$, the number of GAT layers in $\{1, \cdots, 10\}$, the number of hidden dimensions in $\{64, 80, 128, 256\}$, the number of heads in $\{1, 2, 4, 8\}$, dropout in $\{0.1, 0.2, 0.3, 0.5\}$, and learning rate in $\{5e-4, 1e-3\}$. The rest hyperparameters are fixed following the original implementation.

\input{fig-tex/app-relation}
\section{Full Results for Informative Source Nodes}\label{sec:app-full-rel}
To analyze the relations between target and source nodes in the cluster with the highest attention score, we compute the ratio of the source nodes (1) that have the same label as the target nodes, and (2) that are connected with the target nodes, both relative to the total number of source nodes in the cluster. The results for all three backbones are presented in Fig.~\ref{fig:app-rel}. The ratio of source nodes that are connected to their target nodes varies across the different backbones. Among the backbones, SGFormer assigns more attention to source nodes connected with the target nodes than Polynormer and GraphGPS. The label distribution differs between heterophilic and homophilic graphs, yet remains consistent across the backbones. On heterophilic graphs, target nodes assign higher attention scores to source nodes that have the same label as them. This stems from the heterophily of these graphs, where connected nodes often have different labels. Therefore, models only aggregate messages from nodes with different labels via local message passing and turn to global attention to collect messages from nodes with the same label as the target nodes.

\input{fig-tex/app-time}

\section{Complexity Analysis}\label{sec:app-time}
We construct synthetic graphs to test the time cost across different graph scales. The average degree is set to $8$, with a maximum number of edges equals $100,000$. The number of classes is set to $10$. Both the number of input features and hidden dimensions are set to $64$. Following the common configuration of the backbones, we set the number of local layers and global layers as (3, 1), (3, 3), and (6, 3) for SGFormer, GraphGPS, and Polynormer, respectively. All models have $4$ attention heads. From the results in Fig.~\ref{fig:app-time}, we can see that the time cost of Wideformer scales linearly with the number of nodes and remains a small ratio of the total time cost as the number of clusters increases.

In Sec~\ref{sssec:ablation}, we show that increasing the number of clusters $m$ can benefit the model performance, while the optimal value of $m$ is around $3\sim5$. In practice, $m$ is chosen from $\{2,\cdots,8\}$ to remain a small value. Therefore, although a larger $m$ increases the memory and runtime cost linearly, the additional cost is minor. We further present the average cost increase ratio (calculated as $\frac{\mathtt{cost}(m=4)-\mathtt{cost}(m=1)}{\mathtt{cost}(m=1)}$) across different numbers of graph nodes in Tab.~\ref{tab:app-cost-ratio}. We can see that the additional costs are relatively minor, amounting to only $\sim~30\%$ of the original time cost and less than $11\%$ of the memory cost.

\input{tab-tex/app-cost-ratio}
\input{tab-tex/app-enact}

\section{Broader Related Work}
Our study is also related to other works beyond graph transformers. Wideformer employs the center selection method to assign source nodes to different clusters. Traditional cluster pooling methods focus on learning hierarchical features of input graphs~\cite{ying_HierarchicalGraphRepresentation_2018,lee_SelfAttentionGraphPooling_2019,yuan_StructPoolStructuredGraph_2020,mesquita_RethinkingPoolingGraph_2020}. They can collect global information by stacking multiple layers. In contrast, Wideformer and other graph transformer models can collect global information in a single layer, giving rise to shallow architectures for large graphs. 

To better optimize the learning process of the attention matrix with dense interactions, a previous study proposes to inject uniform attention to vision transformers~\cite{hyeon-woo_ScratchingVisualTransformers_2023}. However, this solution cannot be directly adopted for graph transformers. It requires explicitly computing the whole attention matrix, which demands a space complexity of $O(n^2)$ and is not scalable to large graphs. Moreover, different from the occasional observation of the uniform attention~\cite{hyeon-woo_ScratchingVisualTransformers_2023}, we discover that uniform attention with high attention entropy consistently exists across different global graph attention methods on different datasets. To further reduce the computing complexity, ENACT proposes to cluster source nodes during attention-based aggregation for vision transformers~\cite{savathrakis_EnactEntropyBasedClustering_2025}. However, ENACT cannot effectively alleviate over-aggregating. Specifically, their clustering process is based on the features of each individual node, while Wideformer clusters nodes based on the source-target relation, which better enables the selection of the informative sources for the targets. ENACT also applies attention aggregation at the cluster level, while Wideformer applies attention aggregation at the node level within each cluster. Node-level aggregation allows the model to adjust the attention scores at a finer level and capture the node feature distribution within each cluster. To further compare Wideformer against ENACT for tackling over-aggregating, we implement ENACT in graph transformers by directly clustering source nodes based on their own features, and only applying cluster-level attention aggregation. Results in Tab.~\ref{tab:app-enact} show that ENACT cannot effectively tackle over-aggregating and gains worse performance than Wideformer. This demonstrates the effectiveness of our method in tackling over-aggregating and benefiting backbone performance.

\section{Societal Impact}\label{sec:app-impact}
This paper presents work whose goal is to advance the field of 
graph representation learning by proposing a plug-and-play method to tackle over-aggregating problem in graph transformers. The proposed method remains independent of specific downstream applications and can potentially benefit a wide range of domains, such as computational biology~\cite{zaidi_PretrainingDenoisingMolecular_2023, ying_TransformersReallyPerform_2021} and intelligent transportation~\cite{rahmani_GraphNeuralNetworks_2023}. At present, we do not anticipate any evident ethical concerns or foreseeable adverse societal impacts.

\section{Limitation}\label{sec:app-limitation}
This paper discovers the over-aggregating phenomenon in graph attention and proposes a plug-and-play method to tackle over-aggregating. To keep the simplicity of our proposed method, we only explore a simple division method based on KMeans++~\cite{arthur_KmeansAdvantagesCareful_2007} and an attention-guiding method with cluster weighting. In Sec.~\ref{sssec:ablation}, the center selection study primarily shows that a learnable center selection strategy can solve the trade-off between efficiency and model performance. The module ablation study demonstrates that the simple weighting method may fail to benefit backbones under certain circumstances. All these ablation results and the informative source node study in Sec.~\ref{sssec:ana-source} indicates that it lacks further exploration of a learnable division strategy and adaptive weighting method.


\newpage
\section*{NeurIPS Paper Checklist}

\begin{enumerate}

\item {\bf Claims}
    \item[] Question: Do the main claims made in the abstract and introduction accurately reflect the paper's contributions and scope?
    \item[] Answer: \answerYes{} 
    \item[] Justification: The main claims made in the abstract and introduction accurately reflect the paper's contributions and scope.
    \item[] Guidelines:
    \begin{itemize}
        \item The answer NA means that the abstract and introduction do not include the claims made in the paper.
        \item The abstract and/or introduction should clearly state the claims made, including the contributions made in the paper and important assumptions and limitations. A No or NA answer to this question will not be perceived well by the reviewers. 
        \item The claims made should match theoretical and experimental results, and reflect how much the results can be expected to generalize to other settings. 
        \item It is fine to include aspirational goals as motivation as long as it is clear that these goals are not attained by the paper. 
    \end{itemize}

\item {\bf Limitations}
    \item[] Question: Does the paper discuss the limitations of the work performed by the authors?
    \item[] Answer: \answerYes{} 
    \item[] Justification: Please refer to Appendix~\ref{sec:app-limitation}.
    \item[] Guidelines:
    \begin{itemize}
        \item The answer NA means that the paper has no limitation while the answer No means that the paper has limitations, but those are not discussed in the paper. 
        \item The authors are encouraged to create a separate "Limitations" section in their paper.
        \item The paper should point out any strong assumptions and how robust the results are to violations of these assumptions (e.g., independence assumptions, noiseless settings, model well-specification, asymptotic approximations only holding locally). The authors should reflect on how these assumptions might be violated in practice and what the implications would be.
        \item The authors should reflect on the scope of the claims made, e.g., if the approach was only tested on a few datasets or with a few runs. In general, empirical results often depend on implicit assumptions, which should be articulated.
        \item The authors should reflect on the factors that influence the performance of the approach. For example, a facial recognition algorithm may perform poorly when image resolution is low or images are taken in low lighting. Or a speech-to-text system might not be used reliably to provide closed captions for online lectures because it fails to handle technical jargon.
        \item The authors should discuss the computational efficiency of the proposed algorithms and how they scale with dataset size.
        \item If applicable, the authors should discuss possible limitations of their approach to address problems of privacy and fairness.
        \item While the authors might fear that complete honesty about limitations might be used by reviewers as grounds for rejection, a worse outcome might be that reviewers discover limitations that aren't acknowledged in the paper. The authors should use their best judgment and recognize that individual actions in favor of transparency play an important role in developing norms that preserve the integrity of the community. Reviewers will be specifically instructed to not penalize honesty concerning limitations.
    \end{itemize}

\item {\bf Theory assumptions and proofs}
    \item[] Question: For each theoretical result, does the paper provide the full set of assumptions and a complete (and correct) proof?
    \item[] Answer: \answerYes{} 
    \item[] Justification: Theoretical proof is provided for Theorem~\ref{thrm:mono-lower-bound} in Appendix~\ref{sec:app-proof}. Detailed derivation is provided for Eq.~\ref{eq:gradient} in Appendix~\ref{sec:app-derivation}.
    \item[] Guidelines:
    \begin{itemize}
        \item The answer NA means that the paper does not include theoretical results. 
        \item All the theorems, formulas, and proofs in the paper should be numbered and cross-referenced.
        \item All assumptions should be clearly stated or referenced in the statement of any theorems.
        \item The proofs can either appear in the main paper or the supplemental material, but if they appear in the supplemental material, the authors are encouraged to provide a short proof sketch to provide intuition. 
        \item Inversely, any informal proof provided in the core of the paper should be complemented by formal proofs provided in appendix or supplemental material.
        \item Theorems and Lemmas that the proof relies upon should be properly referenced. 
    \end{itemize}

    \item {\bf Experimental result reproducibility}
    \item[] Question: Does the paper fully disclose all the information needed to reproduce the main experimental results of the paper to the extent that it affects the main claims and/or conclusions of the paper (regardless of whether the code and data are provided or not)?
    \item[] Answer: \answerYes{} 
    \item[] Justification: Please refer to Appendix~\ref{ssec:app-setup}. Code is provided in https://github.com/sunjss/over-aggregating.
    \item[] Guidelines:
    \begin{itemize}
        \item The answer NA means that the paper does not include experiments.
        \item If the paper includes experiments, a No answer to this question will not be perceived well by the reviewers: Making the paper reproducible is important, regardless of whether the code and data are provided or not.
        \item If the contribution is a dataset and/or model, the authors should describe the steps taken to make their results reproducible or verifiable. 
        \item Depending on the contribution, reproducibility can be accomplished in various ways. For example, if the contribution is a novel architecture, describing the architecture fully might suffice, or if the contribution is a specific model and empirical evaluation, it may be necessary to either make it possible for others to replicate the model with the same dataset, or provide access to the model. In general. releasing code and data is often one good way to accomplish this, but reproducibility can also be provided via detailed instructions for how to replicate the results, access to a hosted model (e.g., in the case of a large language model), releasing of a model checkpoint, or other means that are appropriate to the research performed.
        \item While NeurIPS does not require releasing code, the conference does require all submissions to provide some reasonable avenue for reproducibility, which may depend on the nature of the contribution. For example
        \begin{enumerate}
            \item If the contribution is primarily a new algorithm, the paper should make it clear how to reproduce that algorithm.
            \item If the contribution is primarily a new model architecture, the paper should describe the architecture clearly and fully.
            \item If the contribution is a new model (e.g., a large language model), then there should either be a way to access this model for reproducing the results or a way to reproduce the model (e.g., with an open-source dataset or instructions for how to construct the dataset).
            \item We recognize that reproducibility may be tricky in some cases, in which case authors are welcome to describe the particular way they provide for reproducibility. In the case of closed-source models, it may be that access to the model is limited in some way (e.g., to registered users), but it should be possible for other researchers to have some path to reproducing or verifying the results.
        \end{enumerate}
    \end{itemize}

\item {\bf Open access to data and code}
    \item[] Question: Does the paper provide open access to the data and code, with sufficient instructions to faithfully reproduce the main experimental results, as described in supplemental material?
    \item[] Answer: \answerYes{} 
    \item[] Justification: Code with instructions for reproduction is provided in https://github.com/sunjss/over-aggregating.
    \item[] Guidelines:
    \begin{itemize}
        \item The answer NA means that paper does not include experiments requiring code.
        \item Please see the NeurIPS code and data submission guidelines (\url{https://nips.cc/public/guides/CodeSubmissionPolicy}) for more details.
        \item While we encourage the release of code and data, we understand that this might not be possible, so “No” is an acceptable answer. Papers cannot be rejected simply for not including code, unless this is central to the contribution (e.g., for a new open-source benchmark).
        \item The instructions should contain the exact command and environment needed to run to reproduce the results. See the NeurIPS code and data submission guidelines (\url{https://nips.cc/public/guides/CodeSubmissionPolicy}) for more details.
        \item The authors should provide instructions on data access and preparation, including how to access the raw data, preprocessed data, intermediate data, and generated data, etc.
        \item The authors should provide scripts to reproduce all experimental results for the new proposed method and baselines. If only a subset of experiments are reproducible, they should state which ones are omitted from the script and why.
        \item At submission time, to preserve anonymity, the authors should release anonymized versions (if applicable).
        \item Providing as much information as possible in supplemental material (appended to the paper) is recommended, but including URLs to data and code is permitted.
    \end{itemize}

\item {\bf Experimental setting/details}
    \item[] Question: Does the paper specify all the training and test details (e.g., data splits, hyperparameters, how they were chosen, type of optimizer, etc.) necessary to understand the results?
    \item[] Answer: \answerYes{} 
    \item[] Justification: Detailed experimental setups are provided in Appendix~\ref{ssec:app-setup}.
    \item[] Guidelines:
    \begin{itemize}
        \item The answer NA means that the paper does not include experiments.
        \item The experimental setting should be presented in the core of the paper to a level of detail that is necessary to appreciate the results and make sense of them.
        \item The full details can be provided either with the code, in appendix, or as supplemental material.
    \end{itemize}

\item {\bf Experiment statistical significance}
    \item[] Question: Does the paper report error bars suitably and correctly defined or other appropriate information about the statistical significance of the experiments?
    \item[] Answer: \answerYes{} 
    \item[] Justification: Please refer to Tab.~\ref{tab:cmp-hom}-Tab.~\ref{tab:abl-module}
    \item[] Guidelines:
    \begin{itemize}
        \item The answer NA means that the paper does not include experiments.
        \item The authors should answer "Yes" if the results are accompanied by error bars, confidence intervals, or statistical significance tests, at least for the experiments that support the main claims of the paper.
        \item The factors of variability that the error bars are capturing should be clearly stated (for example, train/test split, initialization, random drawing of some parameter, or overall run with given experimental conditions).
        \item The method for calculating the error bars should be explained (closed form formula, call to a library function, bootstrap, etc.)
        \item The assumptions made should be given (e.g., Normally distributed errors).
        \item It should be clear whether the error bar is the standard deviation or the standard error of the mean.
        \item It is OK to report 1-sigma error bars, but one should state it. The authors should preferably report a 2-sigma error bar than state that they have a 96\% CI, if the hypothesis of Normality of errors is not verified.
        \item For asymmetric distributions, the authors should be careful not to show in tables or figures symmetric error bars that would yield results that are out of range (e.g. negative error rates).
        \item If error bars are reported in tables or plots, The authors should explain in the text how they were calculated and reference the corresponding figures or tables in the text.
    \end{itemize}

\item {\bf Experiments compute resources}
    \item[] Question: For each experiment, does the paper provide sufficient information on the computer resources (type of compute workers, memory, time of execution) needed to reproduce the experiments?
    \item[] Answer: \answerYes{} 
    \item[] Justification: Please refer to Appendix~\ref{ssec:app-setup}.
    \item[] Guidelines:
    \begin{itemize}
        \item The answer NA means that the paper does not include experiments.
        \item The paper should indicate the type of compute workers CPU or GPU, internal cluster, or cloud provider, including relevant memory and storage.
        \item The paper should provide the amount of compute required for each of the individual experimental runs as well as estimate the total compute. 
        \item The paper should disclose whether the full research project required more compute than the experiments reported in the paper (e.g., preliminary or failed experiments that didn't make it into the paper). 
    \end{itemize}
    
\item {\bf Code of ethics}
    \item[] Question: Does the research conducted in the paper conform, in every respect, with the NeurIPS Code of Ethics \url{https://neurips.cc/public/EthicsGuidelines}?
    \item[] Answer: \answerYes{} 
    \item[] Justification: The research conducted in the paper conforms with the NeurIPS Code of Ethics in every respect.
    \item[] Guidelines:
    \begin{itemize}
        \item The answer NA means that the authors have not reviewed the NeurIPS Code of Ethics.
        \item If the authors answer No, they should explain the special circumstances that require a deviation from the Code of Ethics.
        \item The authors should make sure to preserve anonymity (e.g., if there is a special consideration due to laws or regulations in their jurisdiction).
    \end{itemize}

\item {\bf Broader impacts}
    \item[] Question: Does the paper discuss both potential positive societal impacts and negative societal impacts of the work performed?
    \item[] Answer: \answerYes{} 
    \item[] Justification: Please refer to Appendix~\ref{sec:app-impact}.
    \item[] Guidelines:
    \begin{itemize}
        \item The answer NA means that there is no societal impact of the work performed.
        \item If the authors answer NA or No, they should explain why their work has no societal impact or why the paper does not address societal impact.
        \item Examples of negative societal impacts include potential malicious or unintended uses (e.g., disinformation, generating fake profiles, surveillance), fairness considerations (e.g., deployment of technologies that could make decisions that unfairly impact specific groups), privacy considerations, and security considerations.
        \item The conference expects that many papers will be foundational research and not tied to particular applications, let alone deployments. However, if there is a direct path to any negative applications, the authors should point it out. For example, it is legitimate to point out that an improvement in the quality of generative models could be used to generate deepfakes for disinformation. On the other hand, it is not needed to point out that a generic algorithm for optimizing neural networks could enable people to train models that generate Deepfakes faster.
        \item The authors should consider possible harms that could arise when the technology is being used as intended and functioning correctly, harms that could arise when the technology is being used as intended but gives incorrect results, and harms following from (intentional or unintentional) misuse of the technology.
        \item If there are negative societal impacts, the authors could also discuss possible mitigation strategies (e.g., gated release of models, providing defenses in addition to attacks, mechanisms for monitoring misuse, mechanisms to monitor how a system learns from feedback over time, improving the efficiency and accessibility of ML).
    \end{itemize}
    
\item {\bf Safeguards}
    \item[] Question: Does the paper describe safeguards that have been put in place for responsible release of data or models that have a high risk for misuse (e.g., pretrained language models, image generators, or scraped datasets)?
    \item[] Answer: \answerNA{} 
    \item[] Justification: This paper poses no such risks.
    \item[] Guidelines:
    \begin{itemize}
        \item The answer NA means that the paper poses no such risks.
        \item Released models that have a high risk for misuse or dual-use should be released with necessary safeguards to allow for controlled use of the model, for example by requiring that users adhere to usage guidelines or restrictions to access the model or implementing safety filters. 
        \item Datasets that have been scraped from the Internet could pose safety risks. The authors should describe how they avoided releasing unsafe images.
        \item We recognize that providing effective safeguards is challenging, and many papers do not require this, but we encourage authors to take this into account and make a best faith effort.
    \end{itemize}

\item {\bf Licenses for existing assets}
    \item[] Question: Are the creators or original owners of assets (e.g., code, data, models), used in the paper, properly credited and are the license and terms of use explicitly mentioned and properly respected?
    \item[] Answer: \answerYes{} 
    \item[] Justification: Please refer to Appendix~\ref{ssec:app-dataset}.
    \item[] Guidelines:
    \begin{itemize}
        \item The answer NA means that the paper does not use existing assets.
        \item The authors should cite the original paper that produced the code package or dataset.
        \item The authors should state which version of the asset is used and, if possible, include a URL.
        \item The name of the license (e.g., CC-BY 4.0) should be included for each asset.
        \item For scraped data from a particular source (e.g., website), the copyright and terms of service of that source should be provided.
        \item If assets are released, the license, copyright information, and terms of use in the package should be provided. For popular datasets, \url{paperswithcode.com/datasets} has curated licenses for some datasets. Their licensing guide can help determine the license of a dataset.
        \item For existing datasets that are re-packaged, both the original license and the license of the derived asset (if it has changed) should be provided.
        \item If this information is not available online, the authors are encouraged to reach out to the asset's creators.
    \end{itemize}

\item {\bf New assets}
    \item[] Question: Are new assets introduced in the paper well documented and is the documentation provided alongside the assets?
    \item[] Answer: \answerYes{} 
    \item[] Justification: Please refer to https://github.com/sunjss/over-aggregating.
    \item[] Guidelines:
    \begin{itemize}
        \item The answer NA means that the paper does not release new assets.
        \item Researchers should communicate the details of the dataset/code/model as part of their submissions via structured templates. This includes details about training, license, limitations, etc. 
        \item The paper should discuss whether and how consent was obtained from people whose asset is used.
        \item At submission time, remember to anonymize your assets (if applicable). You can either create an anonymized URL or include an anonymized zip file.
    \end{itemize}

\item {\bf Crowdsourcing and research with human subjects}
    \item[] Question: For crowdsourcing experiments and research with human subjects, does the paper include the full text of instructions given to participants and screenshots, if applicable, as well as details about compensation (if any)? 
    \item[] Answer: \answerNA{} 
    \item[] Justification: This paper does not involve crowdsourcing or research with human subjects.
    \item[] Guidelines:
    \begin{itemize}
        \item The answer NA means that the paper does not involve crowdsourcing nor research with human subjects.
        \item Including this information in the supplemental material is fine, but if the main contribution of the paper involves human subjects, then as much detail as possible should be included in the main paper. 
        \item According to the NeurIPS Code of Ethics, workers involved in data collection, curation, or other labor should be paid at least the minimum wage in the country of the data collector. 
    \end{itemize}

\item {\bf Institutional review board (IRB) approvals or equivalent for research with human subjects}
    \item[] Question: Does the paper describe potential risks incurred by study participants, whether such risks were disclosed to the subjects, and whether Institutional Review Board (IRB) approvals (or an equivalent approval/review based on the requirements of your country or institution) were obtained?
    \item[] Answer: \answerNA{} 
    \item[] Justification: This paper does not involve research with human subjects.
    \item[] Guidelines:
    \begin{itemize}
        \item The answer NA means that the paper does not involve crowdsourcing nor research with human subjects.
        \item Depending on the country in which research is conducted, IRB approval (or equivalent) may be required for any human subjects research. If you obtained IRB approval, you should clearly state this in the paper. 
        \item We recognize that the procedures for this may vary significantly between institutions and locations, and we expect authors to adhere to the NeurIPS Code of Ethics and the guidelines for their institution. 
        \item For initial submissions, do not include any information that would break anonymity (if applicable), such as the institution conducting the review.
    \end{itemize}

\item {\bf Declaration of LLM usage}
    \item[] Question: Does the paper describe the usage of LLMs if it is an important, original, or non-standard component of the core methods in this research? Note that if the LLM is used only for writing, editing, or formatting purposes and does not impact the core methodology, scientific rigorousness, or originality of the research, declaration is not required.
    \item[] Answer: \answerNA{} 
    \item[] Justification: The LLM is used only for writing, editing, or formatting purposes and does not impact the core methodology, scientific rigorousness, or originality of the research.
    \item[] Guidelines:
    \begin{itemize}
        \item The answer NA means that the core method development in this research does not involve LLMs as any important, original, or non-standard components.
        \item Please refer to our LLM policy (\url{https://neurips.cc/Conferences/2025/LLM}) for what should or should not be described.
    \end{itemize}

\end{enumerate}

\end{document}

%% file: fig-tex/teaser.tex
\begin{figure*}[htb]
\centering
\vspace{-0.1in}
\hspace{-0.1in}
\subfigure[Average Attention Entropy]{\label{fig:teaser-all}
\begin{minipage}{0.35\linewidth}
    \centering
    \includegraphics[width=\textwidth]{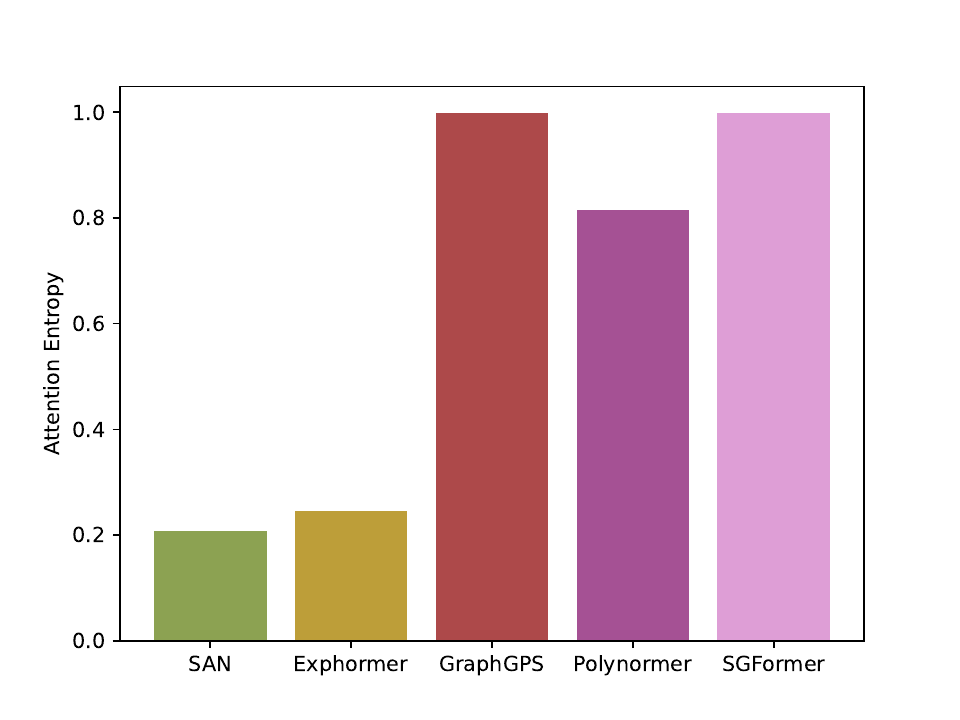}
\end{minipage}}
\hspace{-0.3in}
\subfigure[Entropy across Node Scales]{\label{fig:teaser-node}
\begin{minipage}{0.35\linewidth}
    \centering
    \includegraphics[width=\textwidth]{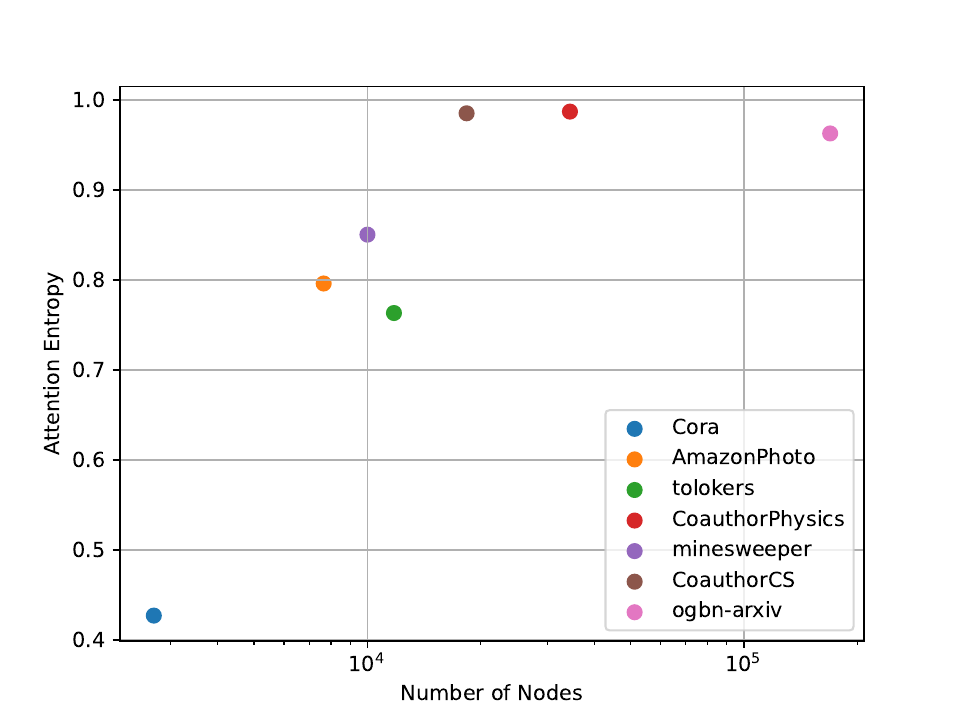}
\end{minipage}}
\hspace{-0.3in}
\subfigure[Entropy during Training]{\label{fig:teaser-train}
\begin{minipage}{0.35\linewidth}
    \centering
    \includegraphics[width=\textwidth]{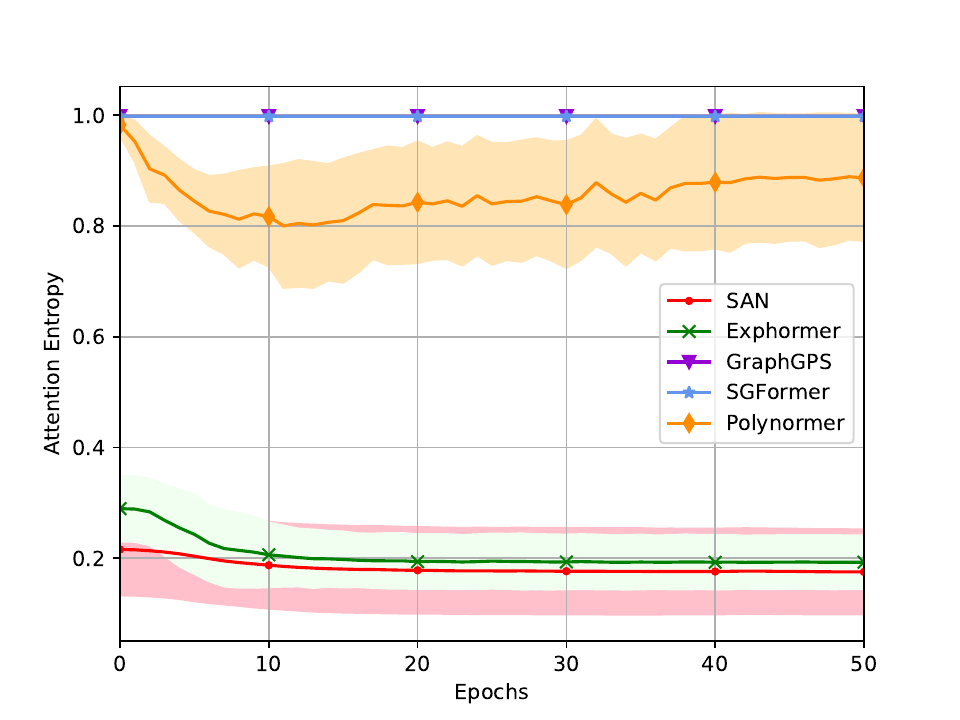}
\end{minipage}}
\caption{\textbf{Over-Aggregating in Graph Transformers.} The entropy values are normalized to $[0,1]$, with higher entropy indicating a more uniform distribution of attention scores. In (b), we show the attention entropy with different numbers of nodes for Polynormer~\cite{deng_PolynormerPolynomialExpressiveGraph_2023}, a typical linear attention method.}
\label{fig:teaser}
\end{figure*}

%% file: tab-tex/teaser.tex
\begin{wrapfigure}{r}{0.42\textwidth}
\vspace{-0.2in}
\tabcaption{\textbf{Effect of Over-Aggregating by Attention Entropy Regularization (Measured by accuracy except for ROC-AUC for minesweeper: \%).}}
\label{tab:teaser}
\begin{center}
\resizebox{0.95\linewidth}{!}{
\begin{small}
\begin{sc}
\begin{tabular}{lcc} 
\toprule
            & Ori        & +Reg                 \\ 
\hline
Cora        & 86.03$_{\color{gray}\pm0.54}$ & \textbf{86.23$_{\color{gray}\pm0.32}$}  \\
Citeseer    & 77.96$_{\color{gray}\pm0.37}$ & \textbf{78.19$_{\color{gray}\pm0.39}$}  \\
AmzPhoto & 95.47$_{\color{gray}\pm0.53}$ & \textbf{95.52$_{\color{gray}\pm0.44}$}  \\
msweeper & 97.13$_{\color{gray}\pm0.17}$ & \textbf{97.20$_{\color{gray}\pm0.32}$}  \\
\bottomrule
\end{tabular}
\end{sc}
\end{small}}
\end{center}
\vspace{-0.1in}
\end{wrapfigure}

%% file: fig-tex/main.tex
\begin{figure*}[htb]
\centering
\includegraphics[width=0.95\textwidth]{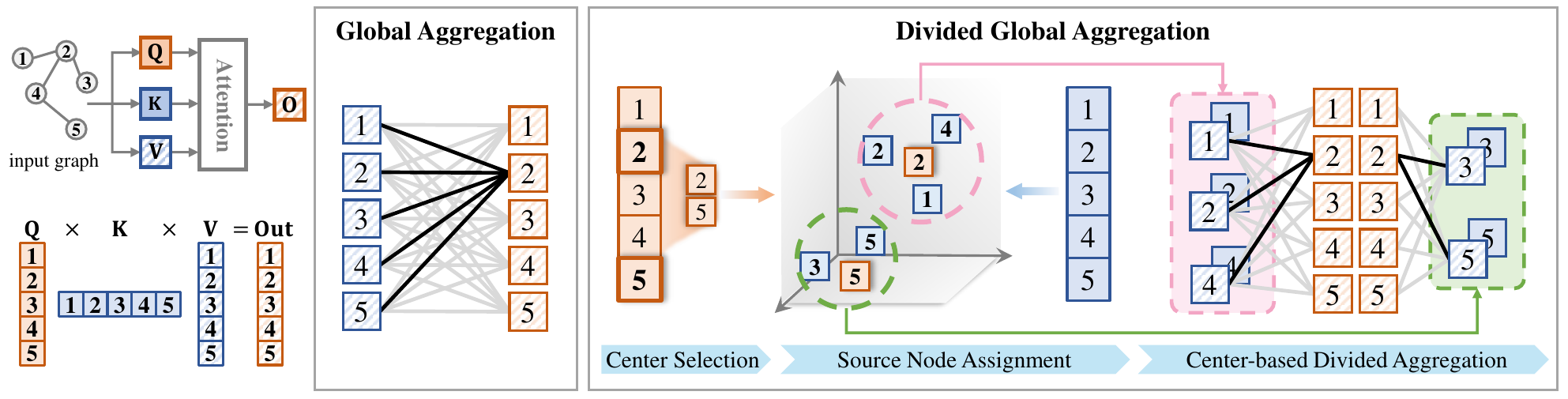}
\vspace{-0.1in}
\caption{\textbf{Attention Comparison between Global Aggregation and Our Divided Global Aggregation.} Global attention employs global aggregation to aggregate source messages into a one-dimensional representation. To relieve the over-aggregating, we assign source nodes into distinct clusters and aggregate each cluster separately, giving rise to multiple aggregation results.}
\vspace{-0.1in}
\label{fig:main}
\end{figure*}

%% file: tab-tex/cmp-hom.tex
\begin{table*}[t]
\caption{\textbf{Comparison Results on Homophilic Datasets (Measured by accuracy: \%).} OOM indicates out-of-memory under the experimental setups.}
\label{tab:cmp-hom}
\begin{center}
\begin{small}
\begin{sc}
\resizebox{0.9\linewidth}{!}{
\begin{tabular}{lcccc} 
\toprule
            & AmazonComputers     & CoauthorCS          & AmazonPhoto         & CoauthorPhysics      \\ 
\hline
\#nodes     & 13,381              & 18,333              & 7,487               & 34,493               \\
\#features  & 767                 & 6,805               & 745                 & 8,415                \\
\#classes   & 10                  & 15                  & 8                   & 5                    \\ 
\hline
GCN         & 89.65$_{\color{gray}\pm0.52}$          & 92.92$_{\color{gray}\pm0.12}$          & 92.70$_{\color{gray}\pm0.20}$          & 96.18$_{\color{gray}\pm0.07}$           \\
GraphSAGE   & 91.20$_{\color{gray}\pm0.29}$          & 93.91$_{\color{gray}\pm0.13}$          & 94.59$_{\color{gray}\pm0.14}$          & 96.49$_{\color{gray}\pm0.06}$           \\
GAT         & 90.78$_{\color{gray}\pm0.13}$          & 93.61$_{\color{gray}\pm0.14}$          & 93.87$_{\color{gray}\pm0.11}$          & 96.17$_{\color{gray}\pm0.08}$           \\
GPRGNN      & 89.32$_{\color{gray}\pm0.29}$          & 95.13$_{\color{gray}\pm0.09}$          & 94.49$_{\color{gray}\pm0.14}$          & 96.85$_{\color{gray}\pm0.08}$           \\ 
\hline
NAGphormer  & 91.19$_{\color{gray}\pm0.14}$          & 95.75$_{\color{gray}\pm0.09}$          & 95.49$_{\color{gray}\pm0.11}$          & 97.34$_{\color{gray}\pm0.03}$           \\
NodeFormer  & 86.98$_{\color{gray}\pm0.62}$          & 95.64$_{\color{gray}\pm0.22}$          & 93.46$_{\color{gray}\pm0.35}$          & 96.45$_{\color{gray}\pm0.28}$           \\
DIFFormer   & 91.99$_{\color{gray}\pm0.76}$          & 94.78$_{\color{gray}\pm0.20}$          & 95.10$_{\color{gray}\pm0.47}$          & 96.60$_{\color{gray}\pm0.18}$           \\
GOAT        & 90.96$_{\color{gray}\pm0.90}$          & 94.21$_{\color{gray}\pm0.38}$          & 92.96$_{\color{gray}\pm1.48}$          & 96.24$_{\color{gray}\pm0.24}$           \\
Exphormer   & 91.47$_{\color{gray}\pm0.17}$          & 94.93$_{\color{gray}\pm0.01}$          & 95.35$_{\color{gray}\pm0.22}$          & 96.89$_{\color{gray}\pm0.09}$           \\ 
\hline
\rowcolor[rgb]{0.894,0.894,0.894}GraphGPS    & 90.62$_{\color{gray}\pm0.36}$          & 95.44$_{\color{gray}\pm0.03}$          & 94.98$_{\color{gray}\pm0.16}$          & 96.75$_{\color{gray}\pm0.07}$           \\
\rowcolor[rgb]{0.894,0.894,0.894}+Entropy Reg    & 90.68$_{\color{gray}\pm0.37}$          & 95.50$_{\color{gray}\pm0.19}$          & 95.28$_{\color{gray}\pm0.28}$          & OOM           \\
\rowcolor[rgb]{0.894,0.894,0.894}+Wideformer & 90.84$_{\color{gray}\pm0.57}$          & \textbf{95.85$_{\color{gray}\pm0.07}$} & 95.55$_{\color{gray}\pm0.74}$          & 96.95$_{\color{gray}\pm0.03}$           \\
SGFormer    & 91.66$_{\color{gray}\pm0.23}$          & 93.50$_{\color{gray}\pm0.32}$          & 95.47$_{\color{gray}\pm0.66}$          & 96.87$_{\color{gray}\pm0.08}$           \\
+Entropy Reg    & 91.78$_{\color{gray}\pm0.37}$          & 93.59$_{\color{gray}\pm0.41}$          & 95.51$_{\color{gray}\pm0.49}$          & 96.88$_{\color{gray}\pm0.09}$           \\
+Wideformer & 92.19$_{\color{gray}\pm0.17}$          & 93.86$_{\color{gray}\pm0.41}$          & 95.64$_{\color{gray}\pm0.57}$          & \textbf{97.39$_{\color{gray}\pm0.19}$}  \\
\rowcolor[rgb]{0.894,0.894,0.894}Polynormer  & 91.98$_{\color{gray}\pm0.09}$          & 94.53$_{\color{gray}\pm0.61}$          & 95.47$_{\color{gray}\pm0.53}$          & 96.67$_{\color{gray}\pm0.29}$           \\
\rowcolor[rgb]{0.894,0.894,0.894}+Entropy Reg  & 92.08$_{\color{gray}\pm0.07}$          & 94.57$_{\color{gray}\pm0.43}$          & 95.52$_{\color{gray}\pm0.44}$          & 96.69$_{\color{gray}\pm0.31}$           \\
\rowcolor[rgb]{0.894,0.894,0.894}+Wideformer & \textbf{92.39$_{\color{gray}\pm0.07}$} & 94.60$_{\color{gray}\pm0.27}$          & \textbf{95.64$_{\color{gray}\pm0.36}$} & 96.79$_{\color{gray}\pm0.20}$           \\
\bottomrule
\end{tabular}
}
\end{sc}
\end{small}
\end{center}
\vspace{-0.2in}
\end{table*}

%% file: fig-tex/over-agg.tex
\begin{wrapfigure}{r}{0.38\linewidth}
\centering
\vspace{-0.8in}
\subfigure[Attention Entropy Comparison]{\label{fig:tackle-oag-all}
\begin{minipage}{\linewidth}
    \centering
    \includegraphics[width=\textwidth]{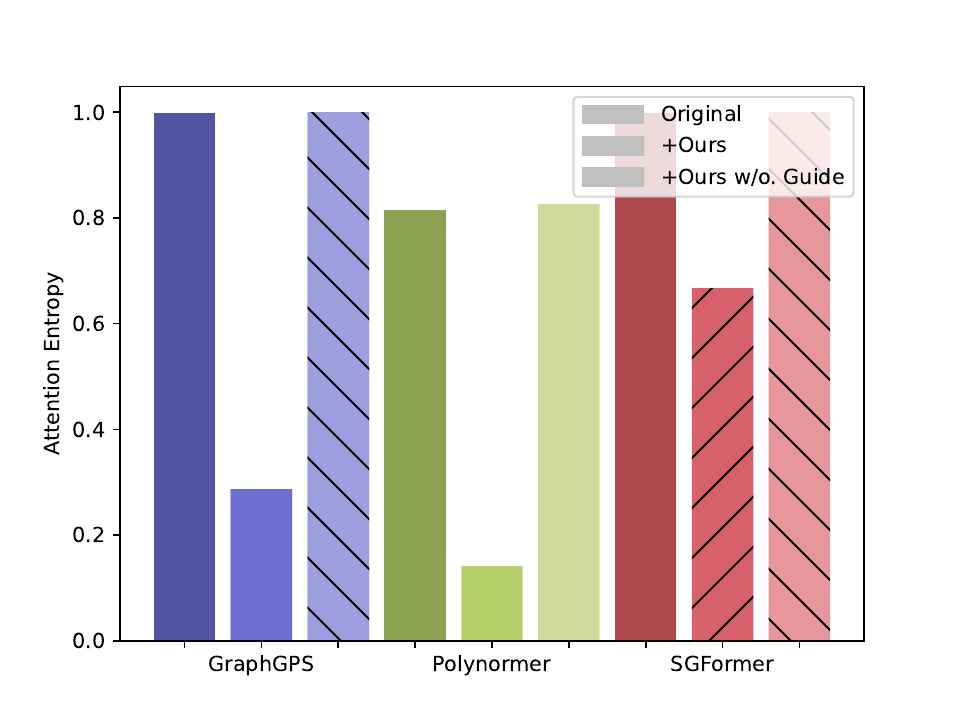}
\end{minipage}}
\subfigure[Cluster Attention Entropy]{\label{fig:tackle-oag-n}
\begin{minipage}{\linewidth}
    \centering
    \vspace{-0.1in}
    \includegraphics[width=\textwidth]{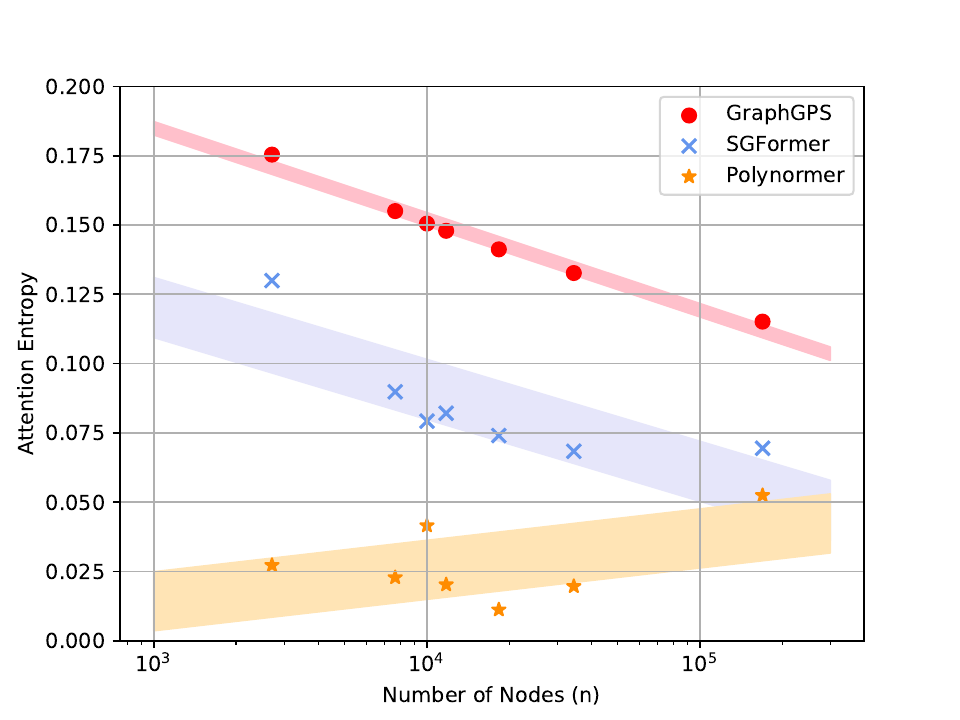}
\end{minipage}}
\caption{\textbf{Effectiveness Evaluation for Over-Aggregating.} ``+ Ours'' denotes using our proposed Wideformer. ``+ Ours w/o. Guide'' employs Wideformer without attention guidance.}
\vspace{-0.4in}
\label{fig:tackle-oag}
\end{wrapfigure}

%% file: tab-tex/cmp-het.tex
\begin{table*}[t]
\caption{\textbf{Comparison Results on Heterophilic Datasets (Measured by ROC-AUC except for accuracy for amazon-ratings and roman-empire: \%).} OOM indicates out-of-memory under the experimental setups.}
\label{tab:cmp-het}
\begin{center}
\begin{small}
\begin{sc}
\resizebox{0.88\linewidth}{!}{
\begin{tabular}{lccccc} 
\toprule
            & amazon-ratings      & minesweeper         & questions           & roman-empire        & tolokers             \\ 
\hline
\#nodes     & 24,492              & 10,000              & 48,921              & 22,662              & 11,758               \\
\#features  & 300                 & 7                   & 301                 & 300                 & 10                   \\
\#classes   & 5                   & 2                   & 2                   & 18                  & 2                    \\ 
\hline
GCN         & 48.70$_{\color{gray}\pm0.63}$          & 89.75$_{\color{gray}\pm0.52}$          & 76.09$_{\color{gray}\pm1.27}$          & 73.69$_{\color{gray}\pm0.74}$          & 83.64$_{\color{gray}\pm0.67}$           \\
GraphSAGE   & 53.63$_{\color{gray}\pm0.39}$          & 93.51$_{\color{gray}\pm0.57}$          & 76.44$_{\color{gray}\pm0.62}$          & 85.74$_{\color{gray}\pm0.67}$          & 82.43$_{\color{gray}\pm0.44}$           \\
GAT         & 52.70$_{\color{gray}\pm0.62}$          & 93.91$_{\color{gray}\pm0.35}$          & 76.79$_{\color{gray}\pm0.71}$          & 88.75$_{\color{gray}\pm0.41}$          & 83.78$_{\color{gray}\pm0.43}$           \\
GPRGNN      & 44.88$_{\color{gray}\pm0.34}$          & 86.24$_{\color{gray}\pm0.61}$          & 55.48$_{\color{gray}\pm0.91}$          & 64.85$_{\color{gray}\pm0.27}$          & 72.94$_{\color{gray}\pm0.97}$           \\ 
\hline
NAGphormer  & 51.26$_{\color{gray}\pm0.72}$          & 84.19$_{\color{gray}\pm0.66}$          & 68.17$_{\color{gray}\pm1.53}$          & 74.34$_{\color{gray}\pm0.77}$          & 78.32$_{\color{gray}\pm0.95}$           \\
NodeFormer  & 43.86$_{\color{gray}\pm0.35}$          & 86.71$_{\color{gray}\pm0.88}$          & 74.27$_{\color{gray}\pm1.46}$          & 64.49$_{\color{gray}\pm0.73}$          & 78.10$_{\color{gray}\pm1.03}$           \\
DIFFormer   & 47.84$_{\color{gray}\pm0.65}$          & 90.89$_{\color{gray}\pm0.58}$          & 72.15$_{\color{gray}\pm1.31}$          & 79.10$_{\color{gray}\pm0.32}$          & 83.57$_{\color{gray}\pm0.68}$           \\
GOAT        & 44.61$_{\color{gray}\pm0.50}$          & 81.09$_{\color{gray}\pm1.02}$          & 75.76$_{\color{gray}\pm1.66}$          & 71.59$_{\color{gray}\pm1.25}$          & 83.11$_{\color{gray}\pm1.04}$           \\
Exphormer   & 53.51$_{\color{gray}\pm0.46}$          & 84.19$_{\color{gray}\pm0.53}$          & 73.94$_{\color{gray}\pm1.06}$          & 89.03$_{\color{gray}\pm0.37}$          & 83.77$_{\color{gray}\pm0.78}$           \\ 
\hline
\rowcolor[rgb]{0.894,0.894,0.894}GraphGPS    & 49.73$_{\color{gray}\pm0.11}$          & 93.26$_{\color{gray}\pm0.10}$          & 75.48$_{\color{gray}\pm0.66}$          & 81.46$_{\color{gray}\pm0.40}$          & 83.95$_{\color{gray}\pm0.81}$           \\
\rowcolor[rgb]{0.894,0.894,0.894}+Entropy Reg    & OOM          & 93.35$_{\color{gray}\pm0.23}$          & OOM          & OOM          & 84.23$_{\color{gray}\pm0.51}$           \\
\rowcolor[rgb]{0.894,0.894,0.894}+Wideformer & 49.96$_{\color{gray}\pm0.41}$          & 93.52$_{\color{gray}\pm0.07}$          & 75.69$_{\color{gray}\pm1.17}$          & 82.12$_{\color{gray}\pm0.39}$          & 84.67$_{\color{gray}\pm0.66}$           \\
SGFormer    & 52.38$_{\color{gray}\pm0.22}$          & 88.60$_{\color{gray}\pm0.49}$          & 76.81$_{\color{gray}\pm0.09}$          & 75.20$_{\color{gray}\pm0.89}$          & 82.24$_{\color{gray}\pm0.13}$           \\
+Entropy Reg    & 52.86$_{\color{gray}\pm0.29}$          & 88.87$_{\color{gray}\pm0.33}$          & 76.83$_{\color{gray}\pm0.18}$          & 75.84$_{\color{gray}\pm0.43}$          & 82.39$_{\color{gray}\pm0.19}$           \\
+Wideformer & 53.47$_{\color{gray}\pm0.14}$          & 89.02$_{\color{gray}\pm0.22}$          & 76.94$_{\color{gray}\pm0.05}$          & 77.03$_{\color{gray}\pm0.38}$          & 82.55$_{\color{gray}\pm0.03}$           \\
\rowcolor[rgb]{0.894,0.894,0.894}Polynormer  & 54.71$_{\color{gray}\pm0.17}$          & 97.13$_{\color{gray}\pm0.17}$          & 78.66$_{\color{gray}\pm0.50}$          & 91.83$_{\color{gray}\pm0.16}$          & 85.09$_{\color{gray}\pm0.21}$           \\
\rowcolor[rgb]{0.894,0.894,0.894}+Entropy Reg  & 54.79$_{\color{gray}\pm0.23}$          & 97.20$_{\color{gray}\pm0.32}$          & 78.69$_{\color{gray}\pm0.24}$          & 91.94$_{\color{gray}\pm0.34}$          & 85.21$_{\color{gray}\pm0.33}$           \\
\rowcolor[rgb]{0.894,0.894,0.894}+Wideformer & \textbf{55.05$_{\color{gray}\pm0.08}$} & \textbf{97.26$_{\color{gray}\pm0.01}$} & \textbf{79.00$_{\color{gray}\pm0.20}$} & \textbf{92.16$_{\color{gray}\pm0.24}$} & \textbf{85.33$_{\color{gray}\pm0.23}$}  \\
\bottomrule
\end{tabular}
}
\end{sc}
\end{small}
\end{center}
\end{table*}

%% file: tab-tex/mix-large+ratio.tex
\begin{table*}
\vspace{-0.1in}
    \begin{minipage}{0.58\linewidth}
        \caption{\textbf{Comparison Results on Datasets with Broader Scales (Measured by accuracy: \%).} OOM indicates out-of-memory under the experimental setups.}\label{tab:cmp-mix}
        \centering
        \resizebox{\textwidth}{!}{
        \input{tab-tex/cmp-mix}}
    \end{minipage}%
    \hfill
    \begin{minipage}{0.38\linewidth}
        \centering
        \includegraphics[width=0.9\linewidth]{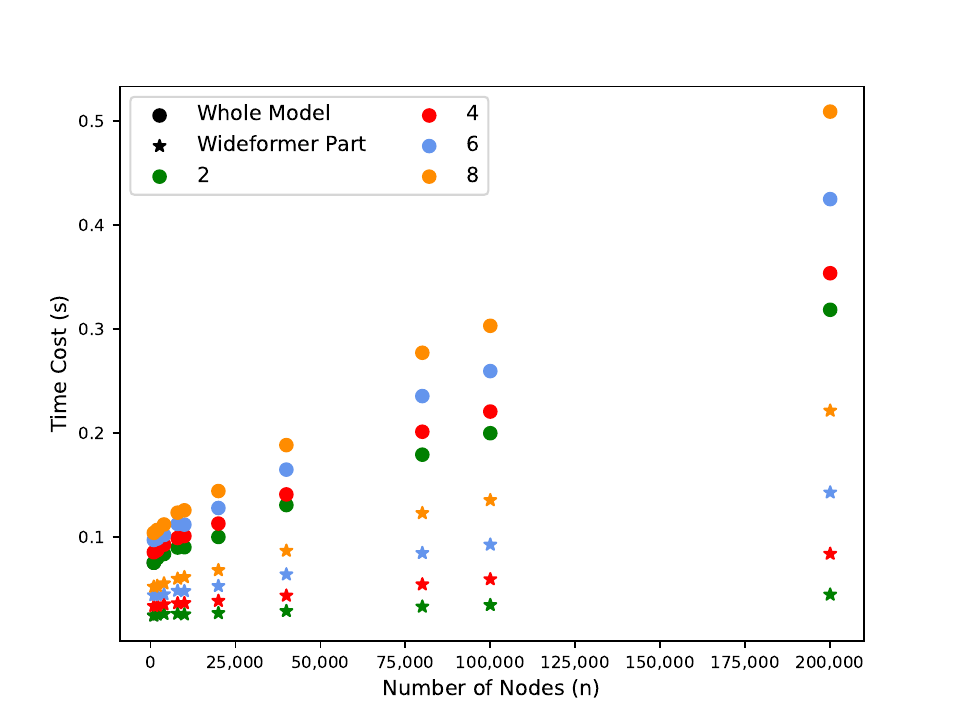}
        \vspace{-0.1in}
        \figcaption{\textbf{Time Cost on GraphGPS.} The point shape ``.'' denotes the whole model and ``*'' denotes the Wideformer part. The color of the points indicates the number of the clusters $m$.}\label{fig:time-gps}
    \end{minipage}
\end{table*}

%% file: tab-tex/cmp-mix.tex
\begin{small}
\begin{sc}
\begin{tabular}{lcccc} 
\toprule
            & ogb-arxiv           & CiteSeer            & Cora                & twitch-gamer  \\ 
\hline
\#nodes     & 169,343             & 3,327               & 2,708               & 168,114       \\
\#features  & 128                 & 3,703               & 1,433               & 7             \\
\#classes   & ~40                 & 6                   & 7                   & 2             \\ 
\hline
GCN         & 71.74$_{\color{gray}\pm0.29}$          & 71.60$_{\color{gray}\pm0.40}$           & 81.60$_{\color{gray}\pm0.40}$           & 62.18$_{\color{gray}\pm0.26}$    \\
GraphSAGE   & 71.49$_{\color{gray}\pm0.27}$          & 71.93$_{\color{gray}\pm0.85}$          & 82.68$_{\color{gray}\pm0.47}$          & 64.37$_{\color{gray}\pm0.39}$    \\
GAT         & 71.95$_{\color{gray}\pm0.36}$          & 72.10$_{\color{gray}\pm1.10}$            & 83.00$_{\color{gray}\pm0.70}$          & 59.89$_{\color{gray}\pm4.12}$    \\
Exphormer   & 72.44$_{\color{gray}\pm0.28}$          & 71.63$_{\color{gray}\pm1.19}$          & 82.77$_{\color{gray}\pm1.38}$          & 64.30$_{\color{gray}\pm0.16}$     \\ 
\hline
\rowcolor[rgb]{0.894,0.894,0.894}GraphGPS    & 70.47$_{\color{gray}\pm1.56}$          & 77.96$_{\color{gray}\pm0.37}$          & 86.03$_{\color{gray}\pm0.54}$          & 64.98$_{\color{gray}\pm0.26}$    \\
\rowcolor[rgb]{0.894,0.894,0.894}+Entropy Reg    & OOM          & 78.19$_{\color{gray}\pm0.39}$          & 86.23$_{\color{gray}\pm0.32}$          & OOM    \\
\rowcolor[rgb]{0.894,0.894,0.894}+Wideformer & 70.66$_{\color{gray}\pm1.29}$          & \textbf{78.61$_{\color{gray}\pm0.35}$} & \textbf{86.70$_{\color{gray}\pm0.83}$} & 65.46$_{\color{gray}\pm0.20}$    \\
SGFormer    & 72.23$_{\color{gray}\pm0.73}$          & 69.93$_{\color{gray}\pm0.31}$          & 80.83$_{\color{gray}\pm0.52}$          & 65.85$_{\color{gray}\pm0.02}$    \\
+Entropy Reg    & OOM          & 70.12$_{\color{gray}\pm0.41}$          & 81.04$_{\color{gray}\pm0.38}$          & OOM    \\
+Wideformer & \textbf{72.58$_{\color{gray}\pm0.39}$} & 70.63$_{\color{gray}\pm0.78}$          & 81.20$_{\color{gray}\pm0.37}$          &               
66.34$_{\color{gray}\pm0.37}$    \\
\rowcolor[rgb]{0.894,0.894,0.894}Polynormer  & 70.77$_{\color{gray}\pm1.16}$          & 66.43$_{\color{gray}\pm1.50}$          & 78.70$_{\color{gray}\pm0.94}$          & 67.15$_{\color{gray}\pm0.03}$        \\
\rowcolor[rgb]{0.894,0.894,0.894}+Entropy Reg  & OOM          & 67.21$_{\color{gray}\pm0.35}$          & 79.28$_{\color{gray}\pm0.51}$          & OOM        \\
\rowcolor[rgb]{0.894,0.894,0.894}+Wideformer & 70.88$_{\color{gray}\pm0.10}$          & 68.10$_{\color{gray}\pm0.42}$          & 79.90$_{\color{gray}\pm0.86}$          & \textbf{67.33$_{\color{gray}\pm0.15}$}        \\
\bottomrule
\end{tabular}
\end{sc}
\end{small}

%% file: fig-tex/ana-node.tex
\begin{wrapfigure}{r}{0.4\linewidth}
\centering
\vspace{-0.9in}
\begin{minipage}[t]{\linewidth}
    \centering
    \includegraphics[width=\textwidth]{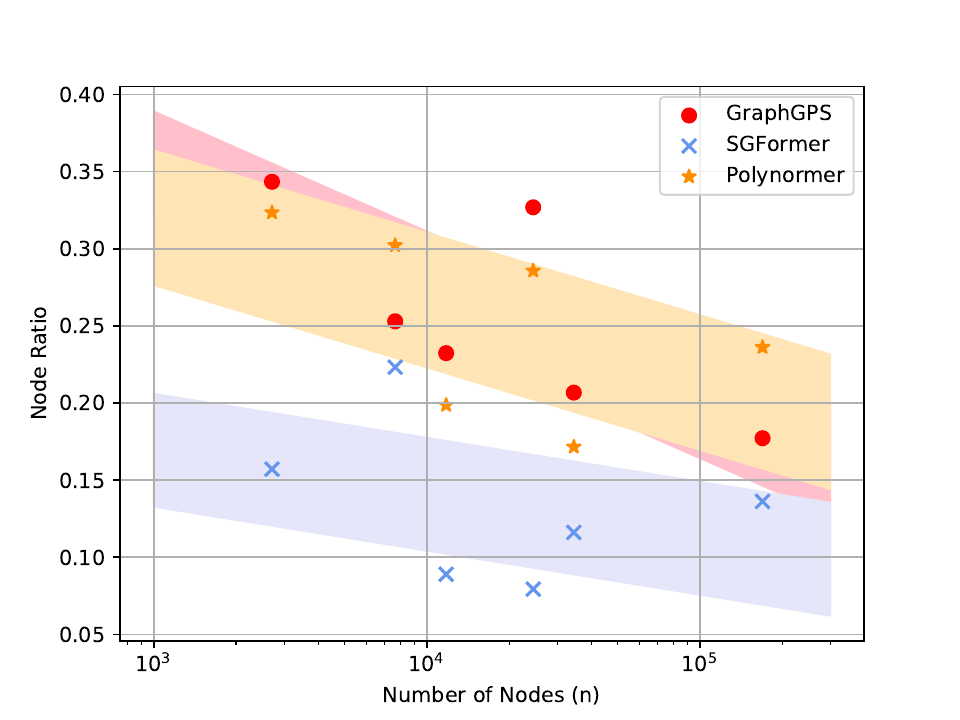}
    \vspace{-0.2in}
    \caption{\textbf{Informative Source Node Ratio.} The ratio is the number of nodes assigned to the cluster with the highest attention score relative to the total number of graph nodes.}\label{fig:att-node-cnt}
\end{minipage}
\hfill
\begin{minipage}[t]{\linewidth}
    \centering
    \includegraphics[width=\textwidth]{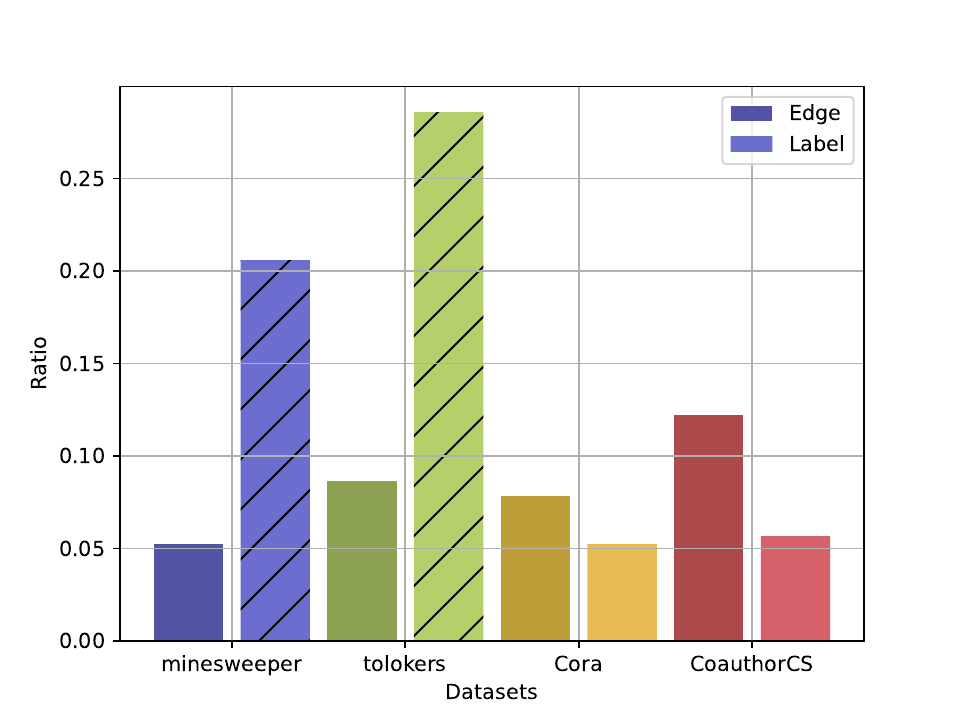}
    \vspace{-0.3in}
    \caption{\textbf{Node Relations.} Label/Edge denotes the ratio of source nodes sharing the same label/connected with the target nodes.}\label{fig:rel-gps}
\end{minipage}
\vspace{-0.5in}
\end{wrapfigure}

%% file: fig-tex/abl-cluster.tex
\begin{wrapfigure}{r}{0.4\linewidth}
\centering
\vspace{-0.3in}
\includegraphics[width=\linewidth]{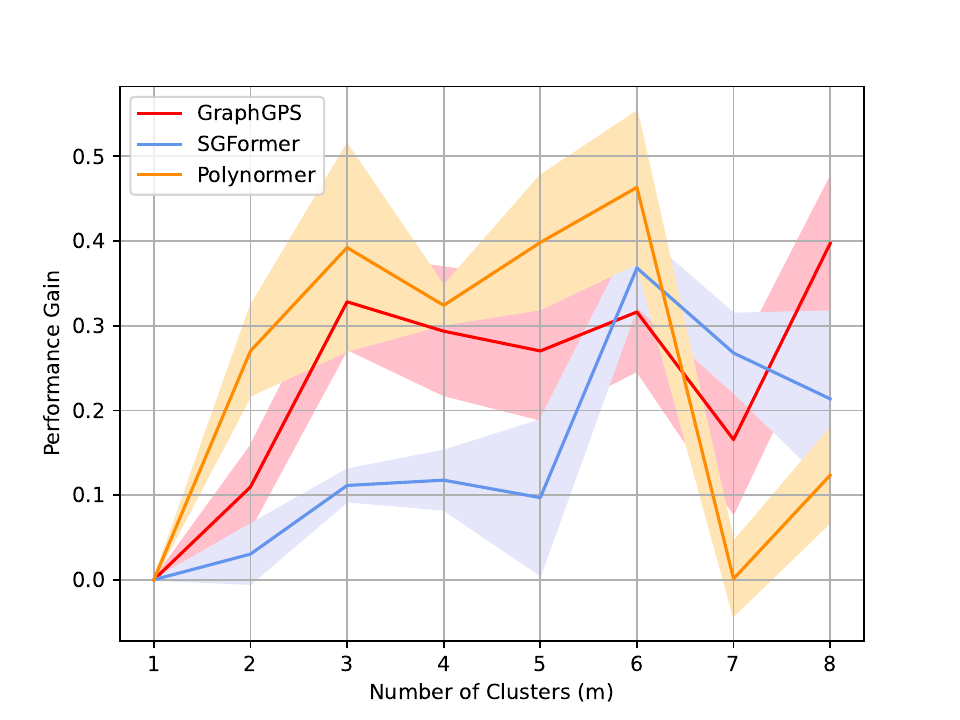}
\vspace{-0.2in}
\caption{\textbf{Comparison with Different Numbers of Clusters.}}\label{fig:abl-cluster}
\vspace{-0.1in}
\end{wrapfigure}

%% file: tab-tex/abl-center.tex
\begin{wrapfigure}{r}{0.45\textwidth}
\vspace{-0.2in}
\tabcaption{\textbf{Comparison on the Center Selection Methods.} Results are averaged on amazon-ratings, questions, CoauthorCS, and AmazonPhoto.}
\label{tab:abl-center}
\begin{center}
\resizebox{\linewidth}{!}{
\begin{small}
\begin{sc}
\begin{tabular}{lccc} 
\toprule
            & GraphGPS            & SGFormer            & Polynormer           \\ 
\hline
Original    & 78.91$_{\color{gray}\pm0.24}$          & 79.54$_{\color{gray}\pm0.32}$          & 80.84$_{\color{gray}\pm0.45}$           \\
+Wideformer & 79.26$_{\color{gray}\pm0.59}$          & 79.98$_{\color{gray}\pm0.29}$          & 81.07$_{\color{gray}\pm0.23}$           \\
+iter=2     & 79.93$_{\color{gray}\pm0.45}$          & 80.58$_{\color{gray}\pm0.39}$          & 81.44$_{\color{gray}\pm0.29}$           \\
+iter=4     & 79.95$_{\color{gray}\pm0.57}$          & 80.42$_{\color{gray}\pm0.37}$          & 81.43$_{\color{gray}\pm0.21}$           \\
+iter=6     & 79.85$_{\color{gray}\pm0.44}$          & 80.55$_{\color{gray}\pm0.34}$          & 81.52$_{\color{gray}\pm0.23}$           \\
+learnable  & \textbf{80.05$_{\color{gray}\pm0.46}$} & \textbf{80.60$_{\color{gray}\pm0.39}$} & \textbf{81.54$_{\color{gray}\pm0.20}$}  \\
\bottomrule
\end{tabular}
\end{sc}
\end{small}}
\end{center}
\vspace{-0.2in}
\end{wrapfigure}

%% file: tab-tex/abl-module.tex
\begin{wrapfigure}{r}{0.45\textwidth}
\vspace{-0.2in}
\tabcaption{\textbf{Module Ablation Results (Measured by accuracy: \%).} ``Div.''/``All'' denotes the divided aggregation without/with attention-guidance.}\label{tab:abl-module}
\vspace{-0.1in}
\begin{center}
\resizebox{\linewidth}{!}{
\begin{small}
\begin{sc}
\begin{tabular}{lcccc} 
\toprule
           & \begin{tabular}[c]{@{}c@{}}amazon\\-ratings\end{tabular} & \begin{tabular}[c]{@{}c@{}}Coauthor\\CS\end{tabular} & \begin{tabular}[c]{@{}c@{}}Amazon\\Photo\end{tabular} & \begin{tabular}[c]{@{}c@{}}roman\\-empire\end{tabular}  \\ 
\hline
GraphGPS   & 49.73$_{\color{gray}\pm0.11}$                                               & 95.44$_{\color{gray}\pm0.03}$                                           & 94.98$_{\color{gray}\pm0.16}$                                            & 81.46$_{\color{gray}\pm0.40}$                                              \\
+Div.      & 49.58$_{\color{gray}\pm0.54}$                                               & 95.56$_{\color{gray}\pm0.30}$                                           & 94.98$_{\color{gray}\pm0.61}$                                            & 81.40$_{\color{gray}\pm0.54}$                                              \\
+All       & \multicolumn{1}{l}{\textbf{\textbf{49.96$_{\color{gray}\pm0.41}$}}}         & \multicolumn{1}{l}{\textbf{\textbf{95.65$_{\color{gray}\pm0.07}$}}}     & \multicolumn{1}{l}{\textbf{\textbf{95.55$_{\color{gray}\pm0.74}$}}}      & \multicolumn{1}{l}{\textbf{\textbf{82.12$_{\color{gray}\pm0.39}$}}}        \\ 
\hline
SGFormer   & 52.38$_{\color{gray}\pm0.22}$                                               & 93.50$_{\color{gray}\pm0.32}$                                           & 95.47$_{\color{gray}\pm0.66}$                                            & 75.20$_{\color{gray}\pm0.89}$                                              \\
+Div.      & 52.42$_{\color{gray}\pm0.54}$                                               & 93.53$_{\color{gray}\pm0.21}$                                           & 95.58$_{\color{gray}\pm0.53}$                                            & \textbf{77.38$_{\color{gray}\pm0.54}$}                                     \\
+All       & \multicolumn{1}{l}{\textbf{\textbf{53.47$_{\color{gray}\pm0.14}$}}}         & \multicolumn{1}{l}{\textbf{\textbf{93.86$_{\color{gray}\pm0.41}$}}}     & \multicolumn{1}{l}{\textbf{\textbf{95.64$_{\color{gray}\pm0.57}$}}}      & \multicolumn{1}{l}{77.03$_{\color{gray}\pm0.38}$}                          \\ 
\hline
Polynormer & 54.71$_{\color{gray}\pm0.17}$                                               & 94.53$_{\color{gray}\pm0.61}$                                           & 95.47$_{\color{gray}\pm0.53}$                                            & 91.83$_{\color{gray}\pm0.16}$                                              \\
+Div.      & \multicolumn{1}{l}{54.82$_{\color{gray}\pm0.11}$}                           & \multicolumn{1}{l}{\textbf{\textbf{94.90$_{\color{gray}\pm0.30}$}}}     & \multicolumn{1}{l}{\textbf{\textbf{96.74$_{\color{gray}\pm0.30}$}}}      & \multicolumn{1}{l}{92.04$_{\color{gray}\pm0.15}$}                          \\
+All       & \textbf{55.05$_{\color{gray}\pm0.08}$}                                      & 94.60$_{\color{gray}\pm0.27}$                                           & 95.64$_{\color{gray}\pm0.36}$                                            & \textbf{92.16$_{\color{gray}\pm0.24}$}                                     \\
\bottomrule
\end{tabular}
\end{sc}
\end{small}}
\end{center}
\vspace{-0.3in}
\end{wrapfigure}

%% file: tab-tex/app-reg.tex
\begin{table*}[t]
\caption{\textbf{Comparison Results with Entropy Regulation on the Cluster Attention (Measure by accuracy: \%).} ``Reg.'' denotes employing the attention entropy of the cluster attention scores as part of the optimization target.}
\label{tab:app-reg}
\begin{center}
\begin{small}
\begin{sc}
\begin{tabular}{lcc} 
\toprule
                    & \multicolumn{1}{l}{roman-empire} & \multicolumn{1}{l}{CoauthorCS}  \\ 
\hline
\rowcolor[rgb]{0.894,0.894,0.894}GraphGPS            & 81.46                            & 95.44                           \\
\rowcolor[rgb]{0.894,0.894,0.894}+Wideformer         & 82.12                            & 95.85                           \\
\rowcolor[rgb]{0.894,0.894,0.894}+Wideformer w/ Reg. & \textbf{82.23}                            & \textbf{96.14}                           \\
SGFormer            & 75.20                            & 93.50                           \\
+Wideformer         & 77.03                            & 93.86                           \\
+Wideformer w/ Reg. & \textbf{77.83}                            & \textbf{94.16}                           \\
\rowcolor[rgb]{0.894,0.894,0.894}Polynormer          & 91.83                            & 94.53                           \\
\rowcolor[rgb]{0.894,0.894,0.894}+Wideformer         & 92.16                            & 94.60                           \\
\rowcolor[rgb]{0.894,0.894,0.894}+Wideformer w/ Reg. & \textbf{92.36}                            & \textbf{95.42}                           \\
\bottomrule
\end{tabular}
\end{sc}
\end{small}
\end{center}
\end{table*}

%% file: fig-tex/app-relation.tex
\begin{figure*}[htb]
\centering
\subfigure[SGFormer]{\label{fig:app-rel-sgformer}
\begin{minipage}{0.31\linewidth}
    \centering
    \includegraphics[width=\textwidth]{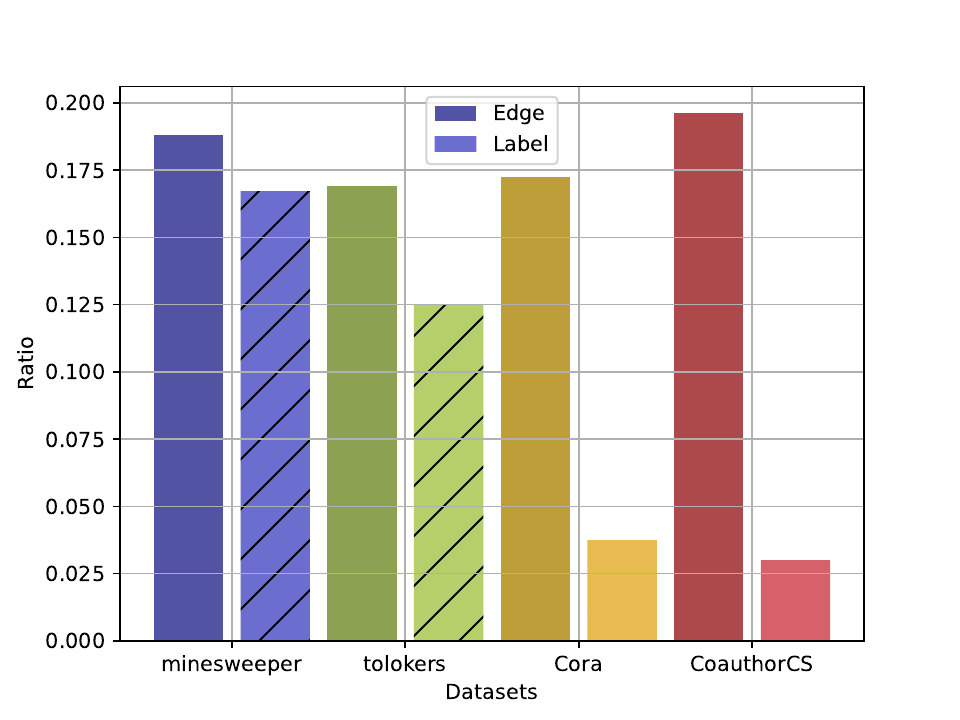}
\end{minipage}}
\hfill
\subfigure[GraphGPS]{\label{fig:app-rel-gps}
\begin{minipage}{0.31\linewidth}
    \centering
    \includegraphics[width=\textwidth]{fig/rel-gps.pdf}
\end{minipage}}
\hfill
\subfigure[Polynormer]{\label{fig:app-rel-poly}
\begin{minipage}{0.31\linewidth}
    \centering
    \includegraphics[width=\textwidth]{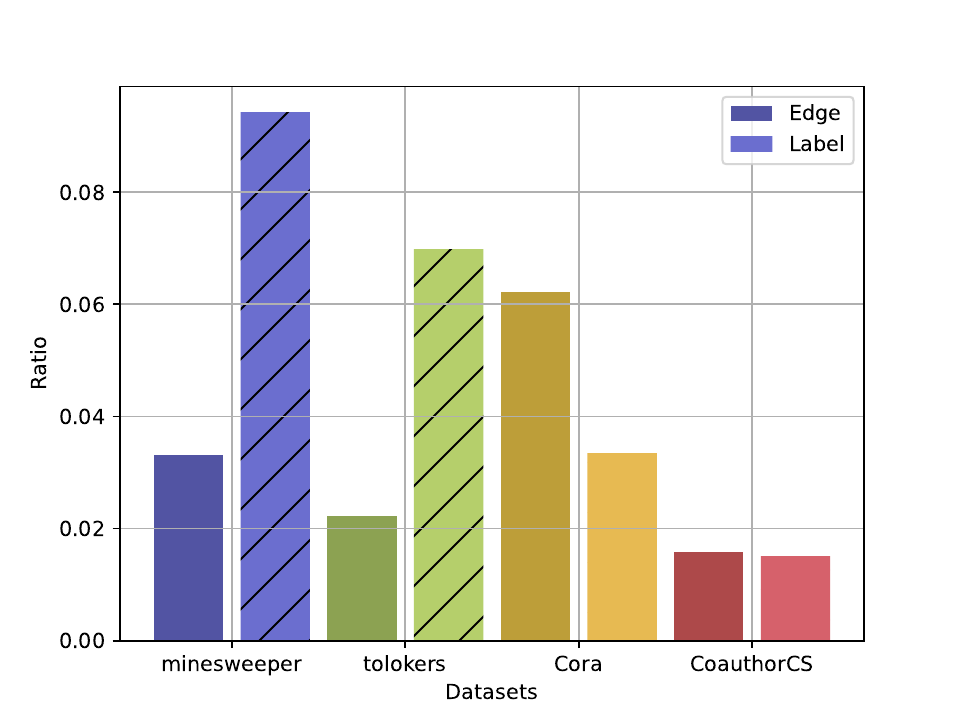}
\end{minipage}}
\caption{\textbf{Full Results for Relations between Source and Target Nodes.}}\label{fig:app-rel}
\end{figure*}

%% file: fig-tex/app-time.tex
\begin{figure*}[htb]
\centering
\subfigure[SGFormer]{\label{fig:app-time-sgformer}
\begin{minipage}{0.32\linewidth}
    \centering
    \includegraphics[width=\textwidth]{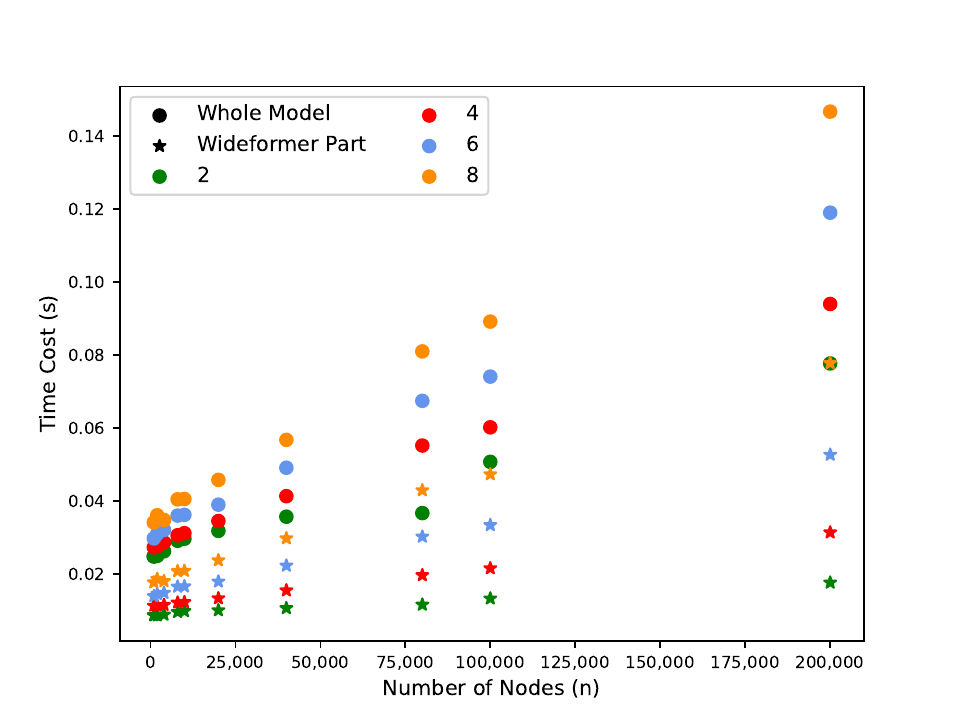}
\end{minipage}}
\subfigure[GraphGPS]{\label{fig:app-time-gps}
\begin{minipage}{0.32\linewidth}
    \centering
    \includegraphics[width=\textwidth]{fig/time-GraphGPS.pdf}
\end{minipage}}
\subfigure[Polynormer]{\label{fig:app-time-polynormer}
\begin{minipage}{0.32\linewidth}
    \centering
    \includegraphics[width=\textwidth]{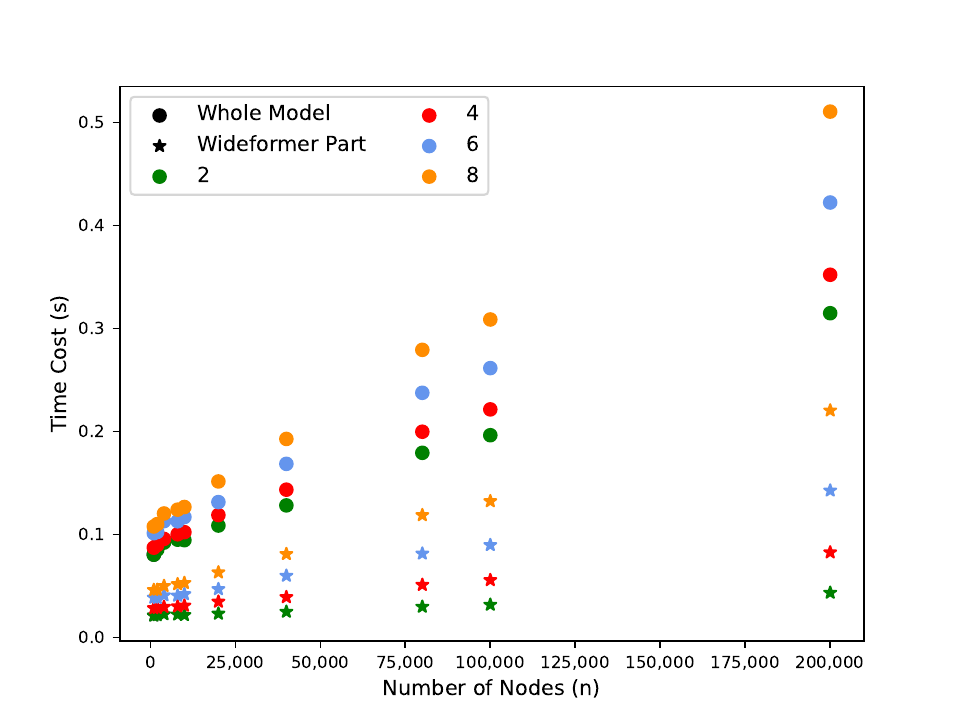}
\end{minipage}}
\caption{\textbf{Time Cost.} Different colors of the points denote the number of clusters $m$. The point shape ``.'' denotes the whole model with Wideformer. ``*'' denotes the Wideformer part.}
\label{fig:app-time}
\end{figure*}

%% file: tab-tex/app-cost-ratio.tex
\begin{table*}[t]
\caption{\textbf{Average Cost Increase Ratio.} Comparing the time and memory cost of 4 clusters against 1 cluster. $m$ denotes the number of clusters. The results are averaged across datasets with different numbers of graph nodes.}
\label{tab:app-cost-ratio}
\begin{center}
\begin{small}
\begin{sc}
\begin{tabular}{lcll} 
\toprule
           & m              & \multicolumn{1}{l}{Time (s)} & \multicolumn{1}{l}{Memory (MB)}  \\ 
\hline
GraphGPS   & 1              & 0.0973                     & 13,870                            \\
           & 4              & 0.1408$_{\textcolor{orange}{30.93\%\uparrow}}$                     & 15,546$_{\textcolor{orange}{10.78\%\uparrow}}$                            \\
\hline
SGFormer   & 1              & 0.0258                      & 50,206                            \\
           & 4              & 0.0413$_{\textcolor{orange}{37.56\%\uparrow}}$                      & 51,332$_{\textcolor{orange}{2.19\%\uparrow}}$                             \\
\hline
Polynormer & 1              & 0.1042                     & 29,338                            \\
           & 4              & 0.1435$_{\textcolor{orange}{27.37\%\uparrow}}$                       & 30,180$_{\textcolor{orange}{2.78\%\uparrow}}$                             \\
\bottomrule
\end{tabular}
\end{sc}
\end{small}
\end{center}
\end{table*}

%% file: tab-tex/app-enact.tex
\begin{table*}[t]
\caption{\textbf{Comparison Results with ENACT (Measure by accuracy except for ROC-AUC for questions: \%).} The bold values denote the best results among comparisons with the same backbone.}
\label{tab:app-enact}
\begin{center}
\begin{small}
\begin{sc}
\begin{tabular}{lcccc} 
\toprule
                                              & amazon-ratings & CoauthorCS     & AmazonPhoto    & questions       \\ 
\hline
\rowcolor[rgb]{0.894,0.894,0.894} GraphGPS    & 49.73          & 95.44          & 94.98          & 75.48           \\
\rowcolor[rgb]{0.894,0.894,0.894} +ENACT      & 48.52          & 95.16          & 94.76          & \textbf{75.74}  \\
\rowcolor[rgb]{0.894,0.894,0.894} +Wideformer & \textbf{49.96} & \textbf{95.85} & \textbf{95.55} & 75.69           \\ 
SGFormer                                      & 52.38          & 93.50          & 95.47          & 76.81           \\
+ENACT                                        & 53.32          & 93.21          & 95.53          & 76.53           \\
+Wideformer                                   & \textbf{53.47} & \textbf{93.86} & \textbf{95.64} & \textbf{76.94}  \\ 
\rowcolor[rgb]{0.894,0.894,0.894} Polynormer  & 54.71          & 94.53          & 95.47          & 78.66           \\
\rowcolor[rgb]{0.894,0.894,0.894} +ENACT      & 54.94          & 94.60          & 95.55          & 78.78           \\
\rowcolor[rgb]{0.894,0.894,0.894} +Wideformer & \textbf{55.05} & \textbf{94.94} & \textbf{95.64} & \textbf{79.00}  \\
\bottomrule
\end{tabular}
\end{sc}
\end{small}
\end{center}
\end{table*}